\documentclass[lettersize,journal]{IEEEtran}
\usepackage{amsmath,amsfonts}
\usepackage{algorithmic}
\usepackage{algorithm}
\usepackage{array}
\usepackage[caption=false]{subfig}
\usepackage{textcomp}
\usepackage{stfloats}
\usepackage{url}
\usepackage{verbatim}
\usepackage{graphicx}
\usepackage{cite}
\usepackage[table]{xcolor}
\usepackage{multirow}
\usepackage{booktabs}
\usepackage{hyperref}

\makeatletter
\long\def\@makecaption#1#2{%
  \footnotesize
  \setbox\@tempboxa\hbox{#1.\ #2}%
  \ifdim \wd\@tempboxa >\hsize
    #1.\ #2\par\vspace{3pt}
  \else
    \hbox to\hsize{\hfil\box\@tempboxa\hfil}\vspace{6pt}
  \fi}
\makeatother

\renewcommand{\baselinestretch}{1.005}

\begin{document}

\title{CRISP: A Spatiotemporal Camera–Radar Backbone for Driving via Forecasting-Based World-Model Pretraining}

\author{%
Jingyu Song$^{*1}$, Yi Liu$^{*1}$, and Katherine A. Skinner$^{1}$
    \thanks{$^*$denotes equal contribution}
    \thanks{$^1$J. Song, Y. Liu, and K. A. Skinner are with the Department of Robotics, University of Michigan, Ann Arbor, MI 48109 USA. Corresponding author e-mail: \tt\small jingyuso@umich.edu}
 }

\markboth{Journal of \LaTeX\ Class Files,~Vol.~14, No.~8, August~2021}%
{Shell \MakeLowercase{\textit{et al.}}: A Sample Article Using IEEEtran.cls for IEEE Journals}

\maketitle

\begin{abstract}
Camera-radar (CR) fusion is a practical sensing configuration for autonomous driving, but existing models are typically trained with task-specific supervision, limiting reusable representation learning. We present CRISP, a spatiotemporal CR backbone pretrained through forecasting-based representation learning. Given historical multi-view images and radar sweeps, CRISP learns a unified bird's-eye-view (BEV) representation by predicting future LiDAR point clouds. LiDAR is used only as privileged supervision during pretraining; the deployed model requires only camera and radar. To make forecasting-based pretraining effective for CR fusion, CRISP introduces an enhanced radar encoder, radar-enhanced temporal self-attention, and multimodal feature rendering with modality innovation gating. These components inject radar range and Doppler cues into BEV temporal propagation and allow BEV tokens to selectively incorporate camera and radar evidence. Experiments on nuScenes show that CRISP improves long-horizon point cloud forecasting and transfers effectively to downstream tasks, including 3D detection, tracking, online mapping, motion forecasting, future occupancy prediction, and planning, suggesting that predictive CR pretraining is a promising path toward scalable driving representations under practical sensor configurations. The project website is \href{https://umfieldrobotics.github.io/CRISP}{https://umfieldrobotics.github.io/CRISP}.
\end{abstract}


\section{Introduction}
\label{sec:ch6-intro}
Autonomous vehicles (AVs) rely on perception systems to understand the surrounding environment and support downstream prediction and planning modules~\cite{nuscenes, sun2020waymo_dataset, hu2023UniAD, wang2023_multi_modal_3d_object_detection_survey, wen2022DL_perception_survey}. Common sensing modalities include camera, LiDAR, and radar, each providing distinct advantages and limitations~\cite{survey_deep_radar, zhao2024crkd}. Cameras offer dense semantic cues, LiDAR provides accurate 3D geometry, and radar is robust to adverse weather and lighting while also measuring motion through Doppler velocity. Since no single sensor is sufficient across all conditions, multimodal fusion has received significant interest from the research community~\cite{survey_detection_multi, survey_deep_radar}.

Among different sensor configurations, LiDAR-camera (LC) fusion has achieved strong performance across many perception benchmarks~\cite{kitti, sun2020waymo_dataset, nuscenes,li2022unifying_uvtr, liu2023bevfusion, FUTR3D}. However, the cost and form factor of LiDAR can limit large-scale deployment in production vehicles~\cite{FUTR3D, zhao2024crkd}. In contrast, camera-radar (CR) sensing offers a more scalable and lower-cost configuration that is already accessible on many vehicles with advanced driver assistance systems~\cite{shi2024radar, zhao2024crkd}. As a result, CR fusion has attracted increasing attention in recent years, with demonstrated benefits for 3D object detection, scene segmentation, mapping, and occupancy prediction~\cite{nabati2021centerfusion, Man2023_BEVGuide, kim2023crn, schramm2024bevcar, ma2024licrocc}. These advances suggest that CR fusion is a promising path toward practical and robust autonomous driving.

\begin{figure}[t]
    \centering
    \includegraphics[width=1.0\linewidth]{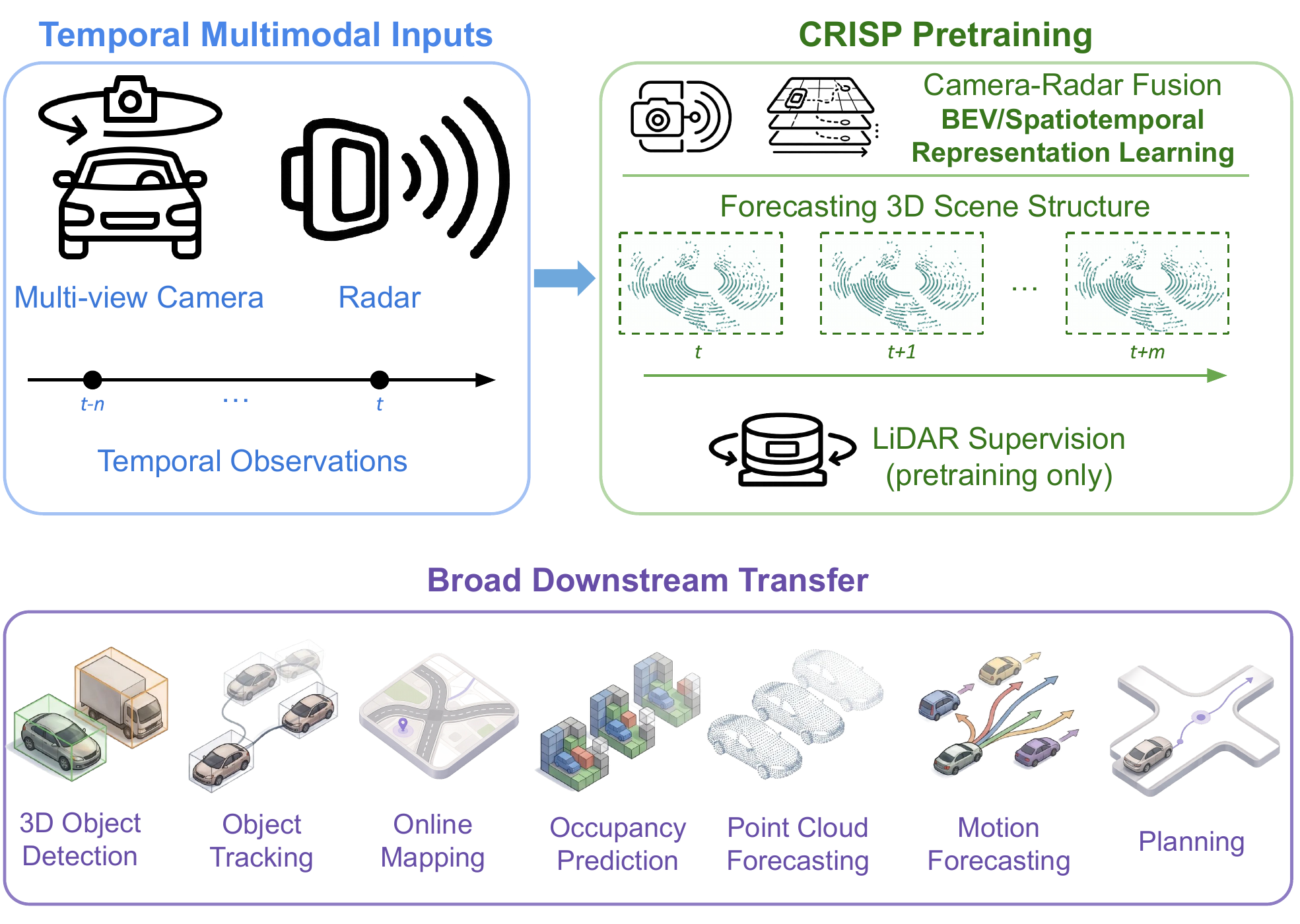}
    \caption{Overview of the motivation of CRISP. We use low-cost camera-radar inputs for inference, while leveraging LiDAR as privileged supervision during pretraining to learn a transferable multimodal driving backbone.}
    \label{fig:pitch}
    \vspace{-2mm}
\end{figure}

Despite this progress, including recent CR fusion networks, many AV perception models~\cite{philion2020lss, li2022bevformer, liu2023bevfusion} still rely heavily on supervised training with precise 3D task annotations. Such annotations are expensive to obtain at scale, which motivates pretraining as a more scalable way to learn effective representations for downstream driving tasks from large amounts of unlabeled or weakly labeled data~\cite{yang2023gd, yang2024unipad, yang2024vidar, wang2025forging}. Here, pretraining denotes an initial representation-learning stage performed before task-specific optimization: rather than training a model from scratch only with downstream labels, the backbone is first optimized with a scalable objective and then finetuned by attaching downstream heads and updating the model with task-specific supervision~\cite{wang2025forging}. Early autonomous driving pretraining methods mainly focus on reconstructing masked LiDAR points, voxels, or bird's-eye-view (BEV) features~\cite{hess2023voxelmae, yang2023gd, lin2024bevmae}, whereas more recent multimodal methods exploit LiDAR as privileged supervision for camera-based or LC pretraining~\cite{park2021pseudo, chen2023pimae, yang2024unipad}. At the same time, forecasting-based pretraining and world-model learning have emerged as effective ways to encode scene dynamics for driving~\cite{khurana2023proxy4docc, min2023uniworld, min2024driveworld, yang2024vidar, wang2025forging}. However, these lines of work are still dominated by camera-only (CO) or LC settings, leaving radar largely unexplored in transferable representation learning.

This gap is particularly important because radar is naturally suited for predictive scene modeling. Cameras provide dense semantic cues, but future motion must be inferred indirectly from appearance, geometry, and temporal changes, and can be affected by depth ambiguity, lighting, and visibility. Radar, in contrast, provides metric range and Doppler velocity measurements, offering a direct physical cue about object motion. These measurements can anchor future scene prediction by complementing visual semantics with instantaneous motion evidence. Despite this suitability for forecasting, existing CR methods are still mostly trained for supervised downstream heads such as detection, segmentation, mapping, or occupancy prediction~\cite{nabati2021centerfusion, kim2023crn, kim2024crt, schramm2024bevcar, ma2024licrocc, zhao2024crkd}. Thus, the synergy between forecasting-based pretraining and CR fusion remains underexplored.

To this end, we propose \textbf{CRISP}, which is summarized in Fig.~\ref{fig:pitch}. CRISP is designed to capture multimodal geometry, radar motion cues, temporal dynamics, and BEV semantics in a unified backbone, while preserving the practical advantage of LiDAR-free inference. Prior forecasting-based pretraining methods model temporal scene evolution effectively, but primarily in CO or LC settings; prior CR methods exploit low-cost multimodal sensing effectively, but usually through task-specific supervised training. Unlike cross-modal alignment or distillation methods that mainly use another modality to regularize same-frame feature learning~\cite{hong2022cross, borse2023x, li2022unifying}, CRISP uses future LiDAR point clouds as privileged forecasting supervision. This encourages the CR backbone to learn not only same-frame geometry, but also multimodal interactions and temporal scene dynamics. LiDAR is used only during pretraining, while the learned backbone remains a practical CR system at inference time. By bridging camera, radar, and LiDAR during pretraining, CRISP provides a scalable representation learning framework to facilitate downstream perception, prediction, and planning tasks.

The main contributions of CRISP are as follows:
\begin{itemize}
    \item We propose CRISP, a forecasting-based CR pretraining framework that learns transferable spatiotemporal BEV representations by predicting future LiDAR point clouds from historical CR observations. LiDAR is used only as privileged supervision during pretraining, while inference remains CR only.

    \item We design a CR spatiotemporal BEV backbone that integrates radar into representation learning through an enhanced radar encoder, radar-enhanced temporal self-attention, and multimodal feature rendering with modality innovation gating. These modules allow radar range and Doppler cues to condition temporal propagation, while enabling each BEV token to selectively compose dense visual semantics with sparse but motion-sensitive radar evidence through gated residual updates.

    \item We demonstrate on nuScenes~\cite{nuscenes} that CRISP improves long-horizon point cloud forecasting and transfers broadly to downstream autonomous driving tasks, including 3D object detection, tracking, online mapping, motion forecasting, future occupancy prediction, and planning, outperforming CO and CR baselines.
\end{itemize}

\section{Related Work}
\label{sec:related_work}

\subsection{Pretraining for Autonomous Driving}

Pretraining has been extensively studied in computer vision and 3D representation learning, and has shown strong transferability across many downstream tasks~\cite{he2022masked, yu2022pointbert, pang2022masked}. Motivated by these advances, recent autonomous driving studies have developed pretraining objectives tailored to outdoor 3D scene understanding~\cite{liang2021gcc3d, hess2023voxelmae, yang2023gd, lin2024bevmae, wang2025forging, visionpad}. These methods typically learn representations by reconstructing masked points, voxels, or BEV features, and have demonstrated that large-scale pretraining can improve downstream 3D perception~\cite{hess2023voxelmae, yang2023gd, lin2024bevmae}. However, most prior works focus on static reconstruction within the current frame, and are therefore less well-suited for learning temporal scene evolution and multimodal interactions jointly.

Multimodal pretraining for autonomous driving has also begun to attract attention~\cite{wang2025forging}. PiMAE~\cite{chen2023pimae} studies interactive masked autoencoding on images and point clouds, while DD3D~\cite{park2021pseudo} and UniPAD~\cite{yang2024unipad} exploit paired LiDAR point clouds to supervise camera-based or multimodal networks through depth prediction or masked scene rendering. These methods demonstrate the value of privileged 3D supervision during pretraining. In parallel, radar-specific pretraining has been explored through contrastive objectives on unlabeled radar data paired with camera observations~\cite{hao2024radical}. Nevertheless, most prior studies either pretrain a single modality or pretrain modalities separately before fusion~\cite{yang2024unipad}, which limits the opportunity to learn fused spatiotemporal representations.

Recent work suggests that forecasting future 3D scene structure provides a stronger pretraining signal for autonomous driving than static reconstruction alone. Point cloud forecasting has been used as a proxy for 4D occupancy forecasting~\cite{khurana2023proxy4docc}, while UniWorld~\cite{min2023uniworld} and DriveWorld~\cite{min2024driveworld} formulate world-model-style pretraining through future occupancy prediction in spatiotemporal scenes. ViDAR~\cite{yang2024vidar} further shows that forecasting future LiDAR point clouds from historical camera observations can improve a broad range of downstream tasks. More recently, LRS4Fusion~\cite{PalladinAndBruckerLRS4Fusion} extends forecasting-based pretraining to sparse LC fusion, demonstrating the promise of pretraining a multimodal backbone for long-range perception.

Despite these advances, existing forecasting-based pretraining methods are still dominated by CO or LC settings. This leaves radar largely unused in transferable representation learning, even though radar provides direct range and Doppler motion measurements that are especially relevant for predicting scene dynamics. LC pretraining benefits from strong metric geometry, but it does not exploit the low-cost motion sensing available in CR systems. In contrast, CRISP targets joint CR pretraining: the backbone learns fused multimodal representations from historical camera and radar observations by forecasting future LiDAR point clouds. This formulation uses LiDAR only as privileged supervision during pretraining, while encouraging the deployed CR backbone to encode cross-modal geometry, radar motion cues, and temporal scene evolution in a unified representation.

\subsection{Camera-Radar Fusion for Autonomous Driving}

CR fusion has gained increasing attention in autonomous driving because the two modalities provide complementary sensing capabilities and improve robustness under adverse weather and lighting conditions~\cite{shi2024radar, wang2019multi, CR_review}. Early CR methods primarily focus on 3D object detection by associating sparse radar measurements with image features. CenterFusion~\cite{nabati2021centerfusion} detects image-space object centers and links them with radar points through frustum association to improve 3D detection and velocity estimation. CramNet~\cite{hwang2022cramnet} further addresses the geometric ambiguity between camera rays and radar returns through ray-constrained cross-attention, and MVFusion~\cite{mvfusion} strengthens cross-modal interactions through semantic alignment and radar-guided transformer fusion. These methods demonstrate the value of radar cues, but they are mostly optimized for detection-specific supervision rather than for learning a general multimodal backbone.

More recently, BEV representations have become a unified spatial framework for integrating camera, LiDAR, and radar features~\cite{philion2020lss, li2022bevformer, liu2023bevfusion, ealss}. Benefiting from the BEV formulation, CR fusion has expanded from object detection to broader driving tasks. RCBEVDet~\cite{lin2024rcbevdet} and CRN~\cite{kim2023crn} learn stronger CR BEV features for 3D perception, while BEV-Guided Fusion~\cite{Man2023_BEVGuide} and BEVCar~\cite{schramm2024bevcar} demonstrate that CR fusion is also beneficial for scene segmentation, mapping, and velocity-aware BEV understanding. In addition, CRT-Fusion~\cite{kim2024crt} highlights the importance of temporal cues for CR-based 3D detection, LiCROcc~\cite{ma2024licrocc} shows that radar can contribute to dense semantic occupancy prediction when guided by LiDAR and camera information, and CRKD~\cite{zhao2024crkd} illustrates the effectiveness of using targeted sensing configurations to improve CR perception through cross-modality knowledge distillation. Most recently, SpaRC-AD~\cite{SpaRC-AD} extends CR fusion to an end-to-end multi-task driving framework, showing that radar can benefit not only detection, but also tracking, motion prediction, and planning.

Despite the rapid progress of CR fusion, most existing methods are still trained with task-specific labels for detection, segmentation, occupancy prediction, or end-to-end driving tasks~\cite{kim2023crn, lin2024rcbevdet, kim2024crt, schramm2024bevcar, ma2024licrocc, zhao2024crkd, SpaRC-AD}. Temporal modeling is typically introduced to improve a particular downstream head, and LiDAR is mainly used through teacher-student supervision or distillation. As a result, it remains unclear how to learn a reusable CR backbone that can support a wide range of downstream autonomous driving tasks. Our work addresses this gap by using forecasting-based pretraining to learn a unified CR backbone before task-specific finetuning.

\subsection{Forecasting-Based World Models for Driving}

Driving world models aim to learn predictive representations of the driving environment: given historical observations, the model should infer the current scene state and anticipate how the 3D scene may evolve in the future. Recent autonomous driving systems increasingly rely on such predictive representations. Temporal BEV encoders such as BEVFormer~\cite{li2022bevformer} show that aggregating historical observations is critical for stable 3D scene understanding, while planning-oriented frameworks such as UniAD~\cite{hu2023UniAD} demonstrate that a shared BEV representation can support perception, motion prediction, and planning in a unified pipeline.

This predictive perspective is closely related to recent driving world models~\cite{wang2025forging}. Point Cloud Forecasting as a Proxy for 4D Occupancy Forecasting~\cite{khurana2023proxy4docc} shows that future 3D scene evolution can be learned from unlabeled sequences through predictive geometry. UniWorld~\cite{min2023uniworld} and DriveWorld~\cite{min2024driveworld} formulate autonomous driving pretraining as spatiotemporal world modeling through future occupancy prediction, while ViDAR~\cite{yang2024vidar} demonstrates that forecasting future LiDAR point clouds from historical camera observations improves downstream perception, prediction, and planning. More recently, HERMES~\cite{zhou2025hermes} proposes a unified driving world model for simultaneous 3D scene understanding and generation, using BEV representations together with large language model (LLM)-based world queries to incorporate contextual world knowledge into future scene prediction. These works suggest that forecasting-based objectives provide a natural training signal for learning representations that encode scene geometry, temporal evolution, and downstream-relevant dynamics.

However, existing predictive world models and forecasting-based pretraining methods mainly focus on visual, LiDAR, or language-augmented inputs, and do not study how radar can be incorporated into a transferable multimodal backbone. This is a notable omission because radar provides direct range and Doppler velocity measurements, which are particularly relevant for predicting dynamic scene evolution. Cameras offer dense semantics but infer motion indirectly, whereas radar provides sparse but physically grounded motion cues. CRISP bridges this gap by using future LiDAR point cloud forecasting as the pretraining objective for a CR backbone. The learned representation is therefore encouraged to encode multimodal geometry, radar motion cues, temporal dynamics, and BEV semantics in a form that transfers not only to perception, but also to downstream prediction and planning tasks~\cite{hu2023UniAD}. This distinguishes CRISP from prior CR fusion networks optimized for supervised task-specific heads, as well as from prior driving world models that do not learn joint CR representations.

\section{Preliminaries}
\label{sec:preliminaries}
We briefly review the two components that CRISP modifies most directly: BEVFormer~\cite{li2022bevformer} for spatiotemporal BEV encoding and ViDAR~\cite{yang2024vidar} for forecasting-based pretraining.

\subsection{BEVFormer}
BEVFormer~\cite{li2022bevformer} is an influential CO framework for
learning BEV representations from multi-view images. Rather
than explicitly constructing a dense 3D volume, BEVFormer uses a set of
grid-shaped BEV queries and transformer attention to aggregate information from
surrounding cameras and past frames. Its unified BEV representation was shown to
support multiple autonomous driving perception tasks, including 3D object
detection and semantic map segmentation, and later became a common BEV encoding
backbone for end-to-end driving systems such as UniAD~\cite{hu2023UniAD}.

Concretely, BEVFormer represents the scene with a grid of learnable BEV queries
and refines them through stacked transformer layers that alternate temporal
self-attention, camera spatial cross-attention (SCA), and feed-forward updates. Given
multi-view image features $\{F_t^i\}_{i=1}^{N_{\text{view}}}$ at time $t$, the BEV encoder aggregates historical context from previous BEV memory and current
spatial evidence from the cameras into a BEV representation.

\subsubsection{Camera Spatial Cross-Attention}
The key view-transformation step in BEVFormer is SCA, which avoids densely
lifting all image features into a 3D volume. For a BEV query
$Q_p$ at grid location $p$, BEVFormer associates the query with a vertical
pillar, samples $N_{\text{ref}}$ 3D reference points along that pillar, projects
these points into each camera view, and attends only to image features from
views in which the projected points are valid:
\begin{equation}
    \mathrm{SCA}(Q_p, F_t)
    =
    \frac{1}{|\mathcal{V}_{\mathrm{hit}}|}
    \sum_{i\in\mathcal{V}_{\mathrm{hit}}}
    \mathrm{DA}\!\left(Q_p,\mathcal{P}_p^i,F_t^i\right),
\end{equation}
where $\mathrm{SCA}(\cdot)$ denotes camera spatial cross-attention,
$Q_p$ is the BEV query at location $p$, and
$F_t=\{F_t^i\}_{i=1}^{N_{\text{view}}}$ denotes the multi-view image
features at time $t$. $\mathcal{P}_p^i$ denotes the image-plane projections,
onto camera $i$, of the $N_{\mathrm{ref}}$ 3D reference points sampled along
the pillar associated with $Q_p$; $\mathcal{V}_{\mathrm{hit}}$ is the set of
camera views where at least one such projected point lies inside the image; and
$\mathrm{DA}$ denotes deformable attention applied to the projected reference
points and the image feature map $F_t^i$. This sparse, geometry-guided attention
makes the BEV encoder camera-aware while avoiding the cost of dense
image-to-volume lifting.

\subsubsection{Standard Temporal Self-Attention (TSA)}
\label{sec:standard-tsa}
BEVFormer incorporates temporal context by attending to ego-motion-compensated
historical BEV memory. Given current BEV queries $Q_t$ and aligned historical
memory $B_{t-1}^{\mathrm{ego}}$, temporal self-attention (TSA) predicts
deformable sampling offsets and attention weights from the temporal BEV queue:
\begin{equation}
\begin{aligned}
    Z_t &= [B_{t-1}^{\mathrm{ego}}, Q_t], \\
    \Delta p_t,\, a_t &= f_{\mathrm{tsa}}(Z_t), \\
    \tilde{B}_t &= \mathrm{MSDeformAttn}(Z_t, p+\Delta p_t, a_t),
\end{aligned}
\end{equation}
where $Z_t=[B_{t-1}^{\mathrm{ego}},Q_t]$ is the temporal BEV queue formed by
concatenating the ego-aligned historical memory and the current query,
$\mathrm{MSDeformAttn}(\cdot)$ denotes multi-scale deformable attention,
$f_{\mathrm{tsa}}(\cdot)$ predicts temporal-attention parameters, $p$ denotes
the BEV reference locations, $\Delta p_t$ denotes the learned sampling offsets,
and $a_t$ denotes the deformable attention weights. In the
standard formulation, the attended features from the temporal queue are fused by
a fixed average, producing the temporally aggregated BEV feature $\bar{B}_t$.

Although ego-motion compensation aligns static scene structure, dynamic objects
may still occupy different BEV locations across frames. Standard TSA must
therefore infer residual object motion implicitly from camera-derived BEV
queries. In CRISP, radar does not need to be injected explicitly into every
temporal operation. Instead, radar information is first blended into the query
state through gated query initialization, producing radar-aware BEV queries.
These queries are then used by the standard temporal attention machinery to
predict sampling offsets, attention weights, and temporal aggregation. As a
result, the offset prediction and temporal fusion become radar-informed through
the query representation itself, while preserving the original TSA interface.

\subsection{ViDAR}

ViDAR~\cite{yang2024vidar} formulates visual autonomous driving pretraining as
future point cloud forecasting: given historical multi-view camera observations,
the model predicts future LiDAR point clouds. This objective encourages the
visual BEV encoder to learn representations that jointly encode semantics, 3D
geometry, and temporal dynamics.

Architecturally, ViDAR builds on BEVFormer~\cite{li2022bevformer} by using it as the history encoder.
The BEVFormer encoder aggregates multi-frame, multi-view camera features into a
spatiotemporal BEV representation, which is then passed through a latent
rendering operator and an autoregressive future prediction decoder. The future
decoder iteratively predicts point clouds at future timestamps from the encoded
historical representation. CRISP keeps this future prediction decoder and the
forecasting objective unchanged, using them as predictive supervision for the
learned BEV features.

The main distinction is the treatment of the intermediate BEV representation.
ViDAR introduces latent rendering to transform camera-derived BEV features into a
3D geometric latent space before future prediction. This is effective for
CO pretraining because the rendering process imposes geometric
structure on features inferred from images. However, the operator is built around
a ray-projection-oriented prior, which is less suitable for further processing
of CR fused features. In CRISP, radar already supplies metric BEV
occupancy and motion cues, and these cues should interact with camera evidence
during BEV formation rather than after a camera-centric rendering step.
Therefore, CRISP preserves ViDAR's predictive decoder but replaces latent
rendering with multimodal feature rendering, where camera and radar information
are composed inside the shared BEV backbone through modality-aware attention and
gated fusion.

\section{Methodology}
\label{sec:methodology}

\begin{figure*}[t]
    \centering
    \includegraphics[width=1.0\linewidth]{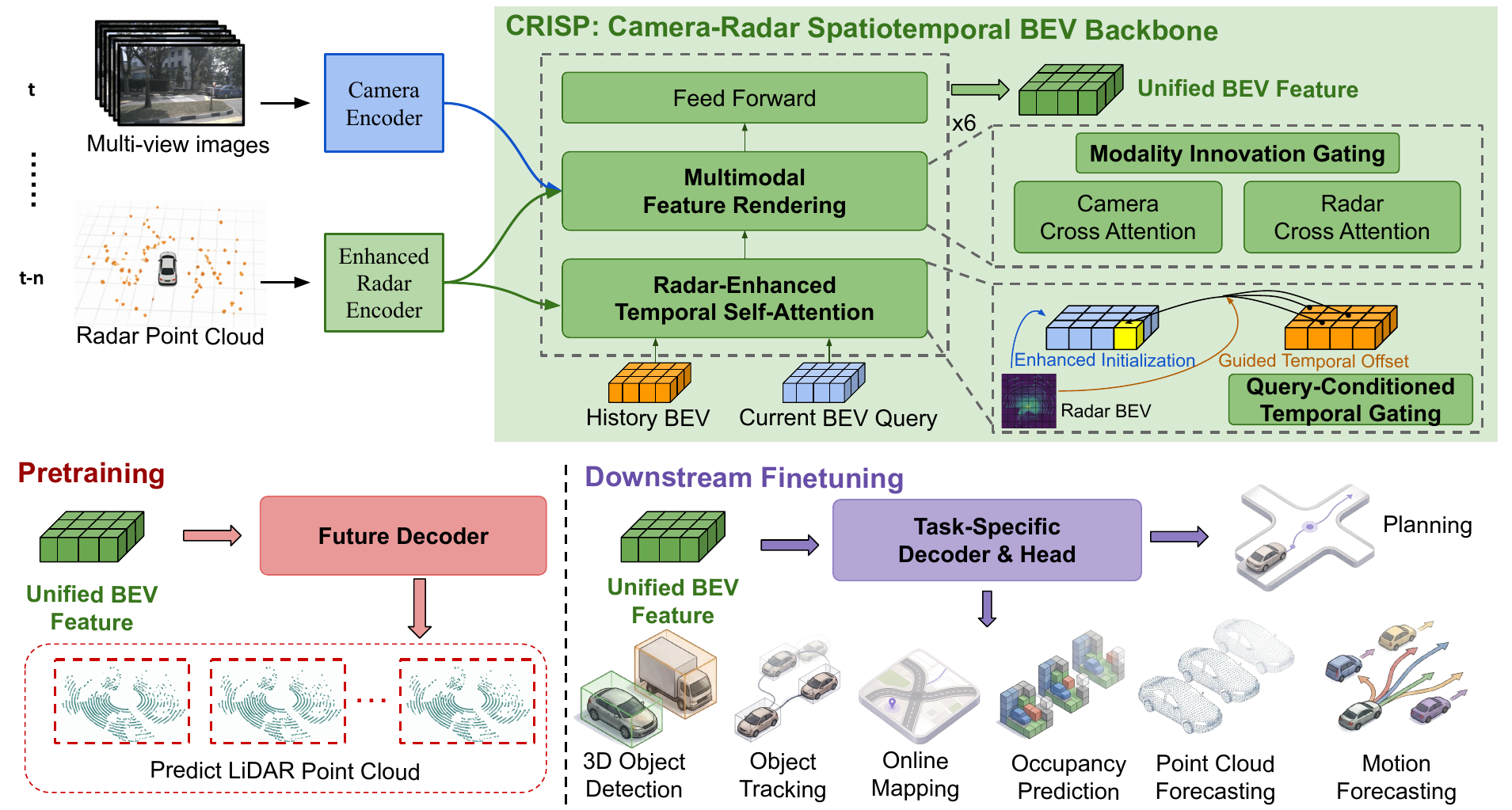}
    \caption{Overview of CRISP. Historical multi-view images and aggregated radar observations are encoded into a shared CR BEV backbone. Radar first passes through ego-aware radar encoding, then enters the backbone through radar-enhanced temporal propagation and multimodal feature rendering. During pretraining, a future occupancy decoder is supervised by future LiDAR; during downstream finetuning, only the pretrained CR backbone is retained.}
    \label{fig:overview}
    \vspace{-4mm}
\end{figure*}

\subsection{Overview of CRISP}

CRISP is a CR spatiotemporal BEV backbone pretrained through future point cloud
forecasting. It learns a joint CR representation directly: cameras contribute
dense visual semantics, while radar contributes sparse metric range and
radial-velocity cues for motion-aware temporal reasoning. Rather than pretraining
a CO BEV encoder and adding radar only during downstream finetuning, CRISP moves
CR interaction inside the BEV backbone and optimizes the resulting representation
with LiDAR-supervised forecasting.

As shown in Fig.~\ref{fig:overview}, CRISP takes historical multi-view images
and radar sweeps as input and encodes them into a unified spatiotemporal BEV
feature. The image stream provides dense visual context, while the radar stream
provides sparse but metric dynamic cues. CRISP fuses these complementary signals
inside the BEV backbone before applying the forecasting objective, so that the
pretrained representation is optimized for both semantic scene understanding and
motion-aware prediction.

The architecture consists of an image encoder, an enhanced radar encoder, a
CR spatiotemporal BEV backbone, and a future prediction decoder used
only during pretraining. The image encoder extracts multi-view image features,
and the radar encoder converts radar sweeps into BEV-aligned radar features.
The shared backbone then produces the unified BEV representation through two
radar-aware operations. First, radar-enhanced temporal self-attention injects
radar information through gated query initialization, making the BEV queries
radar-aware before temporal offsets and attention weights are predicted. Second,
multimodal feature rendering composes camera and radar evidence through
modality-specific cross-attention and adaptive gating.

During pretraining, the unified BEV feature is passed to a future prediction
decoder that predicts current and future scene occupancy supervised by LiDAR point clouds. LiDAR serves only as
privileged supervision and is not required at inference time. During downstream
finetuning, the future decoder is removed and the pretrained CRISP backbone is
connected to task-specific heads for perception, prediction, and planning. This
design uses forecasting to teach the backbone how to combine visual semantics
with radar dynamics, while preserving the benefit of CR-only inference.

\subsection{Enhanced Radar Encoder}

Radar returns provide complementary geometric and motion cues for BEV perception, but their measured Doppler velocities are inherently ego-centric and therefore depend on the instantaneous motion of the sensing vehicle. Directly encoding radar points without this context can make the learned representation ambiguous: similar Doppler patterns may correspond to different object motions under different ego velocities, yaw rates, or accelerations. To address this, we enhance the radar encoder with two lightweight conditioning mechanisms: ego-motion-aware feature modulation and residual spatial calibration. The former injects global vehicle-state information into the radar BEV feature map, while the latter improves feature-level compatibility between radar and camera BEV representations before temporal fusion.

\subsubsection{Ego-Motion Awareness}

Automotive radar measures relative motion with respect to the ego vehicle. Consequently, Doppler-related cues should be interpreted jointly with the current ego state rather than as standalone point attributes. Given the pillar-based radar BEV feature map $F_t^{\mathrm{pill}}$ and the ego-motion information vector $c_t$, we first encode the ego state using a lightweight MLP and broadcast the resulting embedding across the BEV grid:
\begin{equation}
    \tilde{F}_t^{\mathrm{rad}} =
    F_t^{\mathrm{pill}} +
    \mathrm{Broadcast}\!\left(\mathrm{MLP}(c_t)\right).
\end{equation}
This additive conditioning can be interpreted as an ego-motion-dependent feature bias that allows the radar branch to adapt its representation according to the vehicle's instantaneous motion. The conditioned feature map $\tilde{F}_t^{\mathrm{rad}}$ is then passed to the subsequent BEV processing stages, enabling temporal radar features to be formed from ego-aware representations.

\subsubsection{Residual Spatial Calibration}

Although radar and camera features are represented in a common BEV coordinate frame, their feature distributions remain substantially different due to differences in sensing geometry, sparsity, noise characteristics, and semantic content. To reduce this modality gap, we introduce a residual spatial calibrator after ego-motion injection. We first project $\tilde{F}_t^{\mathrm{rad}}$ to a lower-dimensional latent space, concatenate it with a BEV positional encoding $P_{\mathrm{bev}}$, and predict a location-aware correction:
\begin{equation}
    F_t^{\mathrm{rad}} =
    \tilde{F}_t^{\mathrm{rad}} +
    \psi\!\left(
    \mathrm{Proj}\!\left(\tilde{F}_t^{\mathrm{rad}}\right)
    \oplus P_{\mathrm{bev}}
    \right),
\end{equation}
where $\psi(\cdot)$ denotes the calibrator block and $\oplus$ denotes channel-wise concatenation. The residual design preserves the original radar representation while allowing the network to learn spatially varying refinements that improve compatibility with the camera-side BEV features. We use $F_t^{\mathrm{rad}}$ to denote the final ego-motion-aware and BEV-aligned radar feature used throughout the rest of the paper.

\subsection{Camera-Radar Spatiotemporal BEV Backbone}

CRISP extracts unified spatiotemporal BEV features with a CR BEV encoder.
Unlike late-fusion designs that append radar features only after BEV construction, CRISP integrates radar evidence directly inside the BEV encoder, where BEV tokens are temporally propagated and cross-modal evidence is composed. This design is motivated by two observations. First, radar provides geometric and motion-sensitive cues that are useful not only for final detection, but also for deciding how BEV memory should be propagated over time. Second, camera and radar observations have complementary failure modes: camera features are dense and semantically expressive, while radar features are sparse but provide robust range and motion measurements.

CRISP therefore replaces the camera-centric latent rendering stage used in ViDAR with an iterative CR BEV rendering process. Latent rendering is effective for CO pretraining because it encourages image-derived BEV features to acquire geometric structure before future prediction. However, in the CR setting, radar already provides metric BEV observations and motion cues. Passing fused features through a ray-projection-oriented latent rendering operator is therefore less natural than allowing radar and camera evidence to interact directly in the BEV backbone. CRISP preserves ViDAR's future prediction decoder and forecasting objective, but performs multimodal rendering inside the perception encoder.

\begin{figure}[t]
    \centering
    \includegraphics[width=1.0\linewidth]{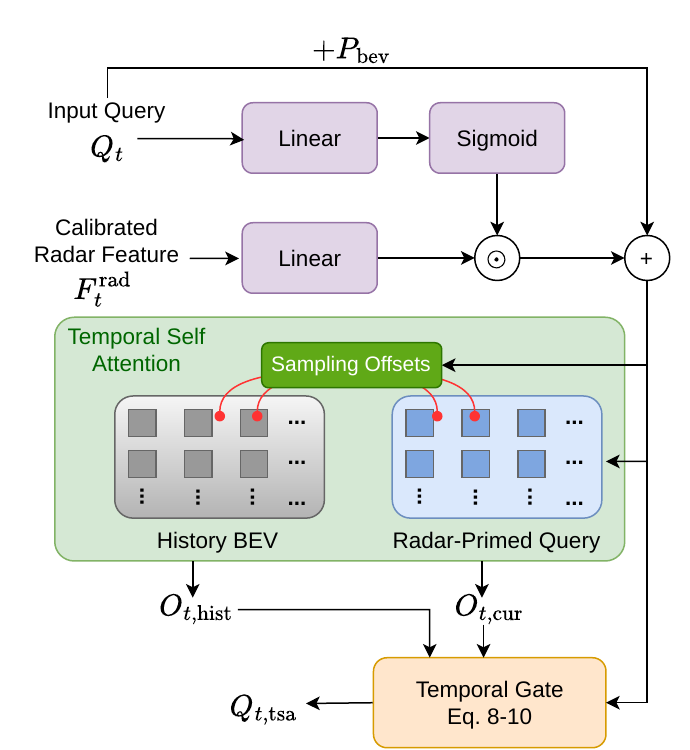}
    \caption{Radar-enhanced temporal self-attention. The calibrated radar BEV feature $F_t^{\mathrm{rad}}$ primes the input BEV query $Q_t^{(l-1)}$ through the query-conditioned gate $g_{t,q}^{(l)}$, producing the radar-primed query $\hat{Q}_t^{(l)}$. Ego-aligned temporal deformable attention then attends to the historical BEV memory $B_{t-1}^{\mathrm{mem}}$ and the current query branch. The temporal gate $g_{t,\mathrm{temp}}^{(l)}$ combines current and historical outputs instead of using uniform averaging.}
    \label{fig:radar_enhanced_tsa}
    \vspace{-4mm}
\end{figure}

The encoder contains $L=6$ repeated layers. Let $Q_t^{(0)} \in \mathbb{R}^{N \times C}$ denote the initial BEV query at timestamp $t$, where $N=H_{\mathrm{bev}} \times W_{\mathrm{bev}}$ and $C$ is the channel dimension. Let $B_{t-1}^{\mathrm{mem}} \in \mathbb{R}^{N \times C}$ denote the ego-motion-aligned BEV memory from the previous frame, let $F_t^{\mathrm{cam}}=\{F_t^i\}_{i=1}^{N_{\text{view}}}$ denote the current multi-view camera features, and let $F_t^{\mathrm{rad}} \in \mathbb{R}^{N \times C}$ denote the calibrated radar BEV feature. Each layer first applies radar-enhanced temporal self-attention, then performs multimodal feature rendering, and finally applies a position-wise feed-forward network (FFN), with residual connections and normalization after each stage. Since every layer has its own attention and fusion parameters, CR fusion is performed iteratively across the backbone. After six layers, the encoder outputs the unified spatiotemporal BEV feature $B_t^{\mathrm{fused}} = Q_t^{(L)}$. We use $B_{t-1}^{\mathrm{mem}}$ for the previous ego-aligned BEV memory consumed by temporal attention, $B_t^{\mathrm{fused}}$ for the current fused BEV state emitted by the backbone, $\mathcal{B}_t$ for the queue of fused BEV states used by the forecasting decoder, and $\hat{B}_{t+\tau}$ for a predicted future BEV state.

A central design choice in CRISP is gated feature update, inspired by recent gated-attention designs~\cite{gated_attention}. Instead of directly adding all attended camera or radar features to the BEV representation, the model predicts gates from the current BEV query and uses them to control each update. We refer to a residual update proposed by an attention branch as an \emph{innovation}: it is new temporal or modality-specific evidence that may modify the current BEV state. In temporal self-attention, the gate balances current-frame features and historical BEV memory. In multimodal feature rendering, the gates control camera-derived and radar-derived innovations.

\subsubsection{Radar-Enhanced Temporal Self-Attention}

As illustrated in Fig.~\ref{fig:radar_enhanced_tsa}, radar-enhanced temporal self-attention introduces radar before temporal aggregation. The calibrated radar BEV feature first modulates the current BEV query through a learned gate, allowing radar range and motion cues to influence temporal attention. Temporal deformable attention then aggregates features from the ego-aligned historical BEV memory and the current radar-primed query. Instead of averaging the two temporal branches (i.e., standard temporal self-attention), CRISP predicts a spatial gate that balances current-frame evidence and propagated history.

For the $l$-th layer, let $Q_t^{(l-1)}$ denote the input BEV query. We first project the calibrated radar feature into the BEV query space and predict a query-conditioned radar gate:
\begin{equation}
    r_t^{(l)}
    =
    \mathrm{Linear}_{r}^{(l)}\!\left(F_t^{\mathrm{rad}}\right),
    \qquad
    g_{t,q}^{(l)}
    =
    \sigma\!\left(
    \mathrm{Linear}_{q}^{(l)}\!\left(Q_t^{(l-1)}\right)
    \right),
\end{equation}
where $r_t^{(l)}$ denotes the projected radar feature used as the residual radar update, $g_{t,q}^{(l)}$ denotes the query-conditioned radar gate, $\sigma(\cdot)$ denotes the sigmoid activation, and $\mathrm{Linear}_{r}^{(l)}(\cdot)$ and $\mathrm{Linear}_{q}^{(l)}(\cdot)$ denote layer-specific affine projections. The current BEV query is then radar-primed through a gated residual update:
\begin{equation}
    \hat{Q}_t^{(l)}
    =
    Q_t^{(l-1)}
    +
    g_{t,q}^{(l)} \odot r_t^{(l)}.
\end{equation}
This \emph{gated radar conditioning} allows radar geometry and motion cues to bias the query used for temporal reasoning, while preserving the original BEV query pathway. The radar-primed query is then combined with the BEV positional encoding:
\begin{equation}
    Q_{t,p}^{(l)}
    =
    \hat{Q}_t^{(l)}
    +
    P_{\mathrm{bev}}.
\end{equation}

The resulting query is used in ego-aligned deformable temporal attention over the previous BEV memory and the current BEV representation. Since the query has already been primed with radar features, the subsequent offset and attention-weight prediction becomes radar-aware through the query representation itself. This corresponds to the guided temporal offset in Fig.~\ref{fig:radar_enhanced_tsa}: the offset predictor is guided by radar because it receives the radar-primed query. This allows radar range and motion cues to benefit temporal sampling naturally.

Let $O_{t,\mathrm{hist}}^{(l)}$ and $O_{t,\mathrm{cur}}^{(l)}$ denote the historical and current outputs of temporal deformable attention. CRISP replaces uniform temporal averaging with a \emph{query-conditioned temporal gate}:
\begin{equation}
    g_{t,\mathrm{temp}}^{(l)}
    =
    \sigma\!\left(
    \phi_{\mathrm{temp}}^{(l)}(Q_{t,p}^{(l)})
    \right).
\end{equation}
Here, $\phi_{\mathrm{temp}}^{(l)}(\cdot)$ denotes a lightweight two-layer MLP. The gated temporal output is
\begin{equation}
    O_{t,\mathrm{temp}}^{(l)}
    =
    g_{t,\mathrm{temp}}^{(l)}
    \odot
    O_{t,\mathrm{cur}}^{(l)}
    +
    \left(1-g_{t,\mathrm{temp}}^{(l)}\right)
    \odot
    O_{t,\mathrm{hist}}^{(l)}.
\end{equation}
Finally, the temporal output is projected back to the BEV query space and added to the radar-primed query:
\begin{equation}
    Q_{t,\mathrm{tsa}}^{(l)}
    =
    \mathrm{LN}\!\left(
    \hat{Q}_t^{(l)}
    +
    \mathrm{Dropout}\!\left(
    \mathrm{Linear}_{o}^{(l)}
    \!\left(
    O_{t,\mathrm{temp}}^{(l)}
    \right)
    \right)
    \right).
\end{equation}
Here, $\mathrm{LN}(\cdot)$ denotes layer normalization.

This module provides a gated temporal update in BEV space. Temporal attention first extracts current and historical features, and the learned gate determines how they are fused. Because this fusion is repeated in every encoder layer, temporal information is progressively refined throughout the six-layer backbone.

\begin{figure}[t]
    \centering
    \includegraphics[width=1.0\linewidth]{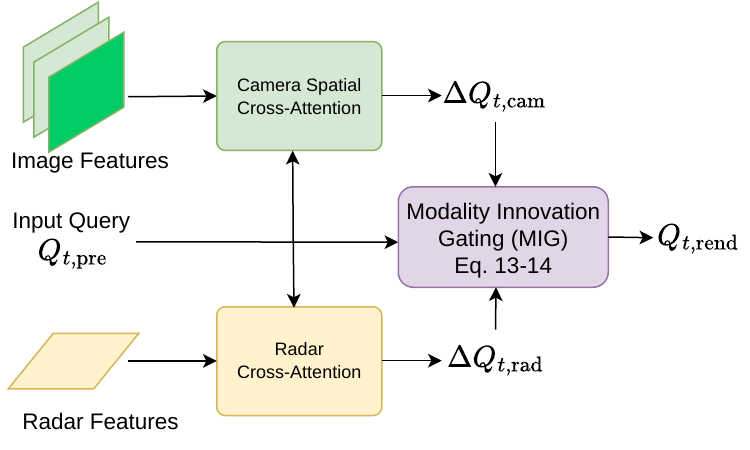}
    \caption{Multimodal feature rendering. Camera and radar cross-attention branches produce residual innovations, which are selectively admitted by the Modality Innovation Gate (MIG). The pre-attention BEV query is preserved through the residual path.}
    \label{fig:multimodal_feature_rendering}
    \vspace{-4mm}
\end{figure}

\subsubsection{Multimodal Feature Rendering}
\label{sec:method-multimodal-rendering}

After temporal aggregation, each BEV token must incorporate sensor evidence from both cameras and radar. As shown in Fig.~\ref{fig:multimodal_feature_rendering}, CRISP replaces ViDAR's camera-centric latent rendering with an in-backbone multimodal rendering step. This choice is important in the CR setting: while latent rendering is useful when geometry must be inferred from images alone, radar already provides calibrated BEV measurements with range and motion cues. Therefore, CRISP directly renders camera and radar evidence into the BEV token at every encoder layer.

We implement this step with a \emph{Modality Innovation Gate} (MIG). The camera and radar branches first produce modality-specific residual updates relative to the current BEV query. MIG predicts separate gates for these updates and controls how much camera and radar information is fused into the BEV representation. Thus, each layer can rely on camera features, radar features, both, or neither, depending on the local scene context.

Let $Q_{t,\mathrm{pre}}^{(l)}=Q_{t,\mathrm{tsa}}^{(l)}$ denote the BEV query before cross-modal interaction. The camera branch uses BEVFormer-style spatial cross-attention to render multi-view image features into the BEV representation:
\begin{equation}
    O_{t,\mathrm{cam}}^{(l)}
    =
    \mathrm{SCA}^{(l)}
    \!\left(
    Q_{t,\mathrm{pre}}^{(l)},
    F_t^{\mathrm{cam}}
    \right),
    \qquad
    \Delta Q_{t,\mathrm{cam}}^{(l)}
    =
    W_{\mathrm{cam}}^{(l)}
    O_{t,\mathrm{cam}}^{(l)}.
\end{equation}
Here, $\mathrm{SCA}(\cdot)$ samples image features from valid camera projections of BEV reference points. In parallel, the radar branch performs BEV-to-BEV deformable radar cross-attention (RCA) over the calibrated radar feature map:
\begin{equation}
    O_{t,\mathrm{rad}}^{(l)}
    =
    \mathrm{RCA}^{(l)}
    \!\left(
    Q_{t,\mathrm{pre}}^{(l)},
    F_t^{\mathrm{rad}}
    \right),
    \qquad
    \Delta Q_{t,\mathrm{rad}}^{(l)}
    =
    W_{\mathrm{rad}}^{(l)}
    O_{t,\mathrm{rad}}^{(l)}.
\end{equation}
Since radar features are already represented in the BEV coordinate frame, $\mathrm{RCA}(\cdot)$ directly samples from the radar BEV feature field using 2D BEV reference points, avoiding the projection and visibility masking required by camera cross-attention.

MIG predicts independent query-conditioned gates for the two modality innovations:
\begin{equation}
    g_{t,\mathrm{cam}}^{(l)}
    =
    \sigma\!\left(
    \phi_{\mathrm{cam}}^{(l)}(Q_{t,\mathrm{pre}}^{(l)})
    \right),
    \qquad
    g_{t,\mathrm{rad}}^{(l)}
    =
    \sigma\!\left(
    \phi_{\mathrm{rad}}^{(l)}(Q_{t,\mathrm{pre}}^{(l)})
    \right).
\end{equation}

Here, $\phi_{\mathrm{cam}}^{(l)}(\cdot)$ and $\phi_{\mathrm{rad}}^{(l)}(\cdot)$ denote layer-specific lightweight MLPs that predict the camera and radar gates, respectively. The gates are implemented group-wise over the channel dimension and are broadcast within each group. For notation simplicity, we write the gated fusion at the token level:
\begin{equation}
    Q_{t,\mathrm{rend}}^{(l)}
    =
    Q_{t,\mathrm{pre}}^{(l)}
    +
    g_{t,\mathrm{cam}}^{(l)}
    \odot
    \Delta Q_{t,\mathrm{cam}}^{(l)}
    +
    g_{t,\mathrm{rad}}^{(l)}
    \odot
    \Delta Q_{t,\mathrm{rad}}^{(l)}.
\end{equation}
The residual path $Q_{t,\mathrm{pre}}^{(l)}$ is preserved explicitly, while MIG controls the camera and radar updates applied to it.

The camera and radar gates are independent rather than competitive (e.g., softmax-normalized). This is important because the two sensors are complementary rather than mutually exclusive. In well-observed regions, both modalities can be useful; for distant or dynamic objects, radar may provide more reliable range and motion cues; for semantic structure, camera features may dominate; and when the existing BEV feature is already confident, both updates can be suppressed. Since each layer has its own MIG parameters, the backbone can adjust the fusion behavior at different depths.

Finally, the rendered feature is passed through a position-wise feed-forward network:
\begin{equation}
    Q_t^{(l)}
    =
    \mathrm{LN}\!\left(
    Q_{t,\mathrm{rend}}^{(l)}
    +
    \mathrm{FFN}^{(l)}
    \!\left(
    Q_{t,\mathrm{rend}}^{(l)}
    \right)
    \right).
\end{equation}
Here, $\mathrm{LN}(\cdot)$ denotes layer normalization.

In this way, multimodal feature rendering extends gated feature update from temporal fusion to sensor fusion. Rather than directly adding all camera and radar attention outputs, CRISP selectively fuses modality updates according to the current BEV representation.

\subsection{Forecasting-Based Pretraining Objective}

CRISP adopts the forecasting-based pretraining objective of ViDAR~\cite{yang2024vidar} to provide predictive geometric supervision for the learned BEV representation. We do not introduce a new forecasting loss; instead, we use the ViDAR-style occupancy prediction decoder to supervise the proposed CR BEV backbone. This allows the backbone to learn geometry, free space, and temporal dynamics from LiDAR point clouds during pretraining, while requiring only camera and radar inputs at inference time.

\subsubsection{Occupancy Prediction Head}

Following ViDAR~\cite{yang2024vidar}, we attach an autoregressive occupancy prediction head only during pretraining. Let $\mathcal{B}_t=\{B_{t-T+1}^{\mathrm{fused}},\ldots,B_t^{\mathrm{fused}}\}$ denote the queue of multimodal BEV states produced by the CR backbone. At future step $\tau$, the decoder predicts a future BEV state conditioned on the previously predicted BEV states and a learned embedding of the relative future ego-motion:
\begin{equation}
    \hat{B}_{t+\tau}
    =
    \Psi_{\mathrm{pred}}\!\left(
    \mathcal{B}_{t+\tau-1},
    e_{\mathrm{ego}}\!\left(\Delta T_{t\rightarrow t+\tau}^{\mathrm{ego}}\right)
    \right),
    \quad
    \tau=1,\ldots,T_{\mathrm{fut}}.
\end{equation}
Here, $\Delta T_{t\rightarrow t+\tau}^{\mathrm{ego}}$ denotes the relative ego-motion from the current frame to the target future frame. During pretraining, it is obtained from the recorded ego trajectory. When the prediction head is used for point cloud forecasting evaluation, we follow ViDAR~\cite{yang2024vidar} and condition all compared methods on the same future ego-motion transform used only to define the future ego-centric coordinate frame. This does not include future sensor observations or future object states.

In our configuration, the prediction decoder follows ViDAR's transformer-based design, but operates on the multimodal BEV representation produced by CRISP. The decoder predicts occupancy logits directly over the BEV grid and vertical bins:
\begin{equation}
    \hat{O}_{t+\tau}
    \in
    \mathbb{R}^{H_{\mathrm{bev}} \times W_{\mathrm{bev}} \times Z},
\end{equation}
where $Z=16$ is the number of vertical occupancy bins. We follow ViDAR~\cite{yang2024vidar} to convert the predicted occupancy volume into a forecast point set for point cloud forecasting evaluation. In addition to future occupancy prediction, we also use auxiliary current-frame occupancy prediction to stabilize training, following ViDAR~\cite{yang2024vidar}.

\subsubsection{Pretraining Loss}

LiDAR point clouds are used as privileged supervision for the predicted occupancy volumes. Following ViDAR~\cite{yang2024vidar}, we supervise the prediction decoder with a sparse LiDAR occupancy classification loss and a dense voxel occupancy reconstruction loss:
\begin{equation}
    \mathcal{L}_{\mathrm{pretrain}}
    =
    \sum_{\tau=0}^{T_{\mathrm{fut}}}
    w_{\tau}
    \left(
    \mathcal{L}_{\mathrm{occ}}^{(\tau)}
    +
    \lambda_{\mathrm{dense}}
    \mathcal{L}_{\mathrm{dense}}^{(\tau)}
    \right),
\end{equation}
where $\tau=0$ denotes the auxiliary current-frame prediction and $\tau>0$ denotes future predictions, $w_{\tau}$ is the loss weight for prediction step $\tau$, and $\lambda_{\mathrm{dense}}$ controls the contribution of the dense voxel loss. The sparse occupancy loss $\mathcal{L}_{\mathrm{occ}}^{(\tau)}$ encourages the predicted occupancy field to assign high probability to LiDAR-observed surface locations and low probability to competing samples along the corresponding sensor line, following the ViDAR formulation. The dense loss $\mathcal{L}_{\mathrm{dense}}^{(\tau)}$ further supervises the reconstructed voxel occupancy field to encourage spatially coherent geometry.
\subsection{Downstream Finetuning}

After pretraining, we discard the occupancy prediction head and retain the camera branch, radar encoder, and CR spatiotemporal BEV backbone as initialization for downstream tasks. Task-specific heads are then attached to the pretrained BEV representation and optimized with the corresponding downstream supervision. The resulting representation is unified, temporally aware, and directly compatible with standard BEV-centric perception, prediction, or planning heads~\cite{li2022bevformer, hu2023UniAD}. Since LiDAR is used only as privileged supervision during pretraining, the finetuned model remains a practical CR system at inference time.

\section{Experiments}
\label{sec:experiments}

\subsection{Experimental Setup}

\subsubsection{Dataset}
We conduct experiments on nuScenes~\cite{nuscenes}, a large-scale autonomous driving dataset with synchronized multi-view cameras, LiDAR, radar, ego-motion, and annotations for perception-centric evaluation. Its synchronized camera--radar streams make it well suited for studying LiDAR-free multimodal representation learning. During pretraining, the model takes historical CR sequences as input, while LiDAR point clouds are used only as privileged supervision for occupancy forecasting. During downstream finetuning and inference, LiDAR is not used; the model operates only with camera and radar inputs.

\subsubsection{Implementation Details}
All main results (Section~\ref{sec:main-results}) use the full CRISP configuration described in Section~\ref{sec:methodology}.
The camera encoder and the spatiotemporal BEV backbone mainly follow common practice in~\cite{li2022bevformer, hu2023UniAD, yang2024vidar}. Specifically, the image backbone (i.e., ResNet-101~\cite{resnet}) is initialized from the checkpoint trained in~\cite{wang2021fcos3d} and a feature pyramid network (FPN)~\cite{fpn} neck is used to produce four levels of image features. The BEV feature dimension is set to $C=256$, and the BEV grid resolution is $200 \times 200$ over the range $[-51.2,51.2]$m in both the longitudinal and lateral directions. The BEV backbone contains six encoder layers.
For radar input, we use 6 radar sweeps. Radar points are voxelized with voxel size $[0.512,0.512,8]$ and encoded by the proposed enhanced radar encoder built on PointPillars~\cite{lang2019pointpillars}. The resulting radar BEV feature is aligned to the same $200\times200$ BEV grid as the camera-side BEV representation. Radar cross-attention is enabled in all six encoder layers, and the Modality Innovation Gate uses 8 channel groups for group-wise gated fusion, matching the number of heads used by deformable attention.

For forecasting-based pretraining, CRISP uses the ViDAR-style occupancy prediction head described in Section~\ref{sec:methodology}. The prediction decoder has three transformer layers and predicts occupancy logits over 16 vertical bins. During training, we supervise the current frame and three future frames using the dense loss weight $\lambda_{\mathrm{dense}}$. The supervised frame weights are set to 1 for the current frame and the three future frames. During forecasting evaluation, the model is rolled out to six future frames.

We pretrain the CRISP backbone for 24 epochs using a learning rate of $2\times10^{-4}$. Experiments are conducted on 8 NVIDIA A100 GPUs with one sample per GPU. After pretraining, we use the pretrained CRISP model as the initialization for downstream models (i.e., BEVFormer~\cite{li2022bevformer} and UniAD~\cite{hu2023UniAD}). Specifically, we retain the camera encoder, radar encoder, and CR spatiotemporal BEV backbone, and connect the resulting unified BEV feature to task-specific heads. Following ViDAR~\cite{yang2024vidar}, we initialize UniAD training with the weights trained for the detection task. We use the same training schedule for BEVFormer and UniAD as ViDAR for a fair comparison.

\subsubsection{Evaluation Protocol}

We evaluate CRISP from two complementary perspectives. First, following~\cite{khurana2023proxy4docc,yang2024vidar}, we directly evaluate point cloud forecasting quality to assess whether the pretrained representation captures future 3D geometry and scene dynamics. We report Chamfer Distance (CD) at future timestamps from 0.5s to 3.0s.

Second, we transfer the pretrained backbone to downstream autonomous driving tasks under the UniAD-style evaluation protocol~\cite{hu2023UniAD, yang2024vidar}. For 3D object detection, we report mean Average Precision (mAP) and nuScenes Detection Score (NDS). For multi-object tracking, we report AMOTA, AMOTP, and Recall, measuring tracking accuracy, localization precision, and object recovery, respectively. For map segmentation, we report intersection-over-union (IoU) for lane, drivable area, divider, and pedestrian crossing categories.
For motion forecasting, we report minimum average displacement error (minADE), minimum final displacement error (minFDE), Miss Rate (MR), and End-to-end Prediction Accuracy (EPA). minADE and minFDE measure the best-mode average and final displacement errors, MR measures the fraction of missed agents, and EPA summarizes end-to-end prediction quality. For future occupancy prediction, we report IoU and Video Panoptic Quality (VPQ) over near and far regions, where IoU measures voxel overlap and VPQ measures spatiotemporal occupancy quality. For planning, we report average $\mathcal{L}_2$ displacement error and average collision rate over the 3s planning horizon. Together, these metrics evaluate whether the pretrained CRISP backbone transfers to perception, prediction, and planning.

\begin{table}[t]
    \centering
    \caption{Point cloud forecasting on nuScenes following the ViDAR protocol~\cite{yang2024vidar}. We report Chamfer Distance (CD) for future timestamps from 0.5s to 3.0s. The best result is shown in \textbf{bold} and the second-best result is \underline{underlined}. $^\dagger$ indicates a language-augmented world model~\cite{zhou2025hermes} with an LLM-based component and additional text supervision, which is different from the sensor-only pretraining setting of the other listed methods.}
    \label{tab:pc_forecasting}
    \setlength{\tabcolsep}{4pt}
    \renewcommand{\arraystretch}{1.05}
    \small
    \resizebox{0.49\textwidth}{!}{%
    \begin{tabular}{c l c c c c c c c}
        \toprule
        \multicolumn{1}{c}{\multirow{2}{*}{\shortstack{History\\Horizon}}} &
        \multicolumn{1}{c}{\multirow{2}{*}{Method}} &
        \multirow{2}{*}{Modality} &
        \multicolumn{6}{c}{Chamfer Distance ($\mathrm{m}^2$) $\downarrow$} \\
        \cmidrule(lr){4-9}
        & & & 0.5s & 1.0s & 1.5s & 2.0s & 2.5s & 3.0s \\
        \midrule
        \multirow{3}{*}{0s}
        & HERMES$^\dagger$~\cite{zhou2025hermes} & C
        & -- & 0.78 & -- & 0.95 & -- & 1.17 \\
        & LRS4Fusion~\cite{PalladinAndBruckerLRS4Fusion} & C+L
        & \textbf{0.35} & \textbf{0.57} & 0.75 & 0.96 & 1.21 & 1.51 \\
        \rowcolor{green!5}
        & CRISP (ours) & C+R
        & \underline{0.52} & \underline{0.59} & \textbf{0.67} & \textbf{0.74} & \textbf{0.82} & \textbf{0.91} \\
        \midrule
        \multirow{4}{*}{1s}
        & 4D-Occ~\cite{khurana2023proxy4docc} & L
        & 1.26 & 1.88 & -- & -- & -- & -- \\
        & ViDAR~\cite{yang2024vidar} & C
        & 1.11 & 1.25 & 1.40 & 1.57 & 1.76 & 1.97 \\
        & LRS4Fusion~\cite{PalladinAndBruckerLRS4Fusion} & C+L
        & \textbf{0.31} & \textbf{0.48} & 0.64 & 0.79 & 0.99 & 1.25 \\
        \rowcolor{green!5}
        & CRISP (ours) & C+R
        & \underline{0.47} & \underline{0.53} & \textbf{0.60} & \textbf{0.68} & \textbf{0.76} & \textbf{0.86} \\
        \midrule
        \multirow{4}{*}{3s}
        & 4D-Occ~\cite{khurana2023proxy4docc} & L
        & 0.91 & 1.13 & 1.30 & 1.53 & 1.72 & 2.11 \\
        & ViDAR~\cite{yang2024vidar} & C
        & 1.01 & 1.12 & 1.25 & 1.38 & 1.54 & 1.73 \\
        & LRS4Fusion~\cite{PalladinAndBruckerLRS4Fusion} & C+L
        & \textbf{0.33} & \textbf{0.47} & 0.61 & 0.77 & 0.97 & 1.23 \\
        \rowcolor{green!5}
        & CRISP (ours) & C+R
        & \underline{0.44} & \underline{0.51} & \textbf{0.58} & \textbf{0.65} & \textbf{0.74} & \textbf{0.84} \\
        \bottomrule
    \end{tabular}
    }
\end{table}

\subsection{Main Results}
\label{sec:main-results}

We evaluate CRISP on point cloud forecasting and a broad set of downstream autonomous driving tasks, including 3D detection, tracking, map segmentation, motion forecasting, future occupancy prediction, and planning. We report three variants of our method:
\begin{itemize}
    \item \textbf{CRISP}: the pretrained CR backbone evaluated with the occupancy forecasting head for future point cloud prediction.
    \item \textbf{BEVFormer-CRISP}: BEVFormer with its original detection head retained and its BEV backbone replaced by the pretrained CRISP backbone.
    \item \textbf{UniAD-CRISP}: UniAD with its original task heads retained and its BEV backbone replaced by the pretrained CRISP backbone.
\end{itemize}

\subsubsection{Point Cloud Forecasting}

Table~\ref{tab:pc_forecasting} evaluates predictive 3D geometry under the ViDAR point cloud forecasting protocol~\cite{yang2024vidar}. ViDAR establishes this setting by pretraining a CO BEV encoder to forecast future LiDAR point clouds, providing a strong baseline for vision-based predictive representation learning. CRISP extends this paradigm to CR pretraining and demonstrates a clear advantage in long-horizon forecasting. While LRS4Fusion~\cite{PalladinAndBruckerLRS4Fusion} benefits from LC inputs and performs best at short horizons, CRISP consistently overtakes it from 1.5s onward across all history settings. This result is particularly notable because LiDAR typically provides much stronger metric geometry than radar; nevertheless, the proposed CR backbone achieves more accurate long-horizon scene evolution. The trend suggests that CRISP does not simply append radar as an auxiliary cue, but uses radar-enhanced temporal attention and multimodal feature rendering to learn motion-aware BEV representations for future prediction. Under the 0s history setting, CRISP also outperforms HERMES~\cite{zhou2025hermes} at the reported horizons, even though HERMES uses an LLM-based language-augmented world-model component and additional text supervision. Overall, these results highlight the central benefit of CRISP: accurate long-horizon 3D forecasting from a practical CR system compatible with LiDAR-free inference.

\subsubsection{Downstream Finetuning}

We next study whether the backbone pretrained through forecasting transfers to downstream autonomous driving tasks. Our main CO baselines are BEVFormer~\cite{li2022bevformer} and UniAD~\cite{hu2023UniAD}, together with their ViDAR-pretrained counterparts. We also compare with SpaRC-AD~\cite{SpaRC-AD}, a strong CR multi-task model. For CRISP transfer, we replace the BEV backbone of BEVFormer or UniAD with the pretrained CRISP backbone while keeping the original downstream heads unchanged.

\begin{table}[t]
    \centering
    \caption{3D detection results. All metrics are reported in percentage. The best result is shown in \textbf{bold} and the second-best result is \underline{underlined}.}
    \label{tab:main-detection}
    \setlength{\tabcolsep}{6pt}
    \renewcommand{\arraystretch}{1.05}
    \small
    \resizebox{0.47\textwidth}{!}{%
    \begin{tabular}{l c c c c}
        \toprule
        Method & Modality & Pretrain & mAP $\uparrow$ & NDS $\uparrow$\\
        \midrule
        \multirow{3}{*}{UVTR~\cite{li2022unifying_uvtr}}
            & \multirow{3}{*}{C}
            & - & 39.2 & 48.8 \\
            &  & UniPAD~\cite{yang2024unipad} & 42.8 & 50.2 \\
            &  & VisionPAD~\cite{visionpad} & 43.1 & 50.4 \\
        \midrule
        \multirow{2}{*}{BEVFormer~\cite{li2022bevformer}}
            & \multirow{2}{*}{C}
            & - & 41.5 & 51.7 \\
            &  & ViDAR~\cite{yang2024vidar} & 45.8 & 54.8 \\
        \midrule
        SpaRC-AD~\cite{SpaRC-AD}
            & C+R
            & -
            & 46.6 & 57.0 \\
        \midrule
        \multirow{2}{*}{BEVFormer-CRISP}
            & \multirow{2}{*}{C+R}
            & -
            & \underline{49.8} & \underline{59.1} \\
            &
            & \cellcolor{green!5}CRISP
            & \cellcolor{green!5}\textbf{53.2}
            & \cellcolor{green!5}\textbf{61.3} \\
        \bottomrule
    \end{tabular}
    }
\end{table}

\begin{table}[t]
    \centering
    \caption{Online map segmentation results. All IoU values are percentages. The best result is shown in \textbf{bold} and the second-best result is \underline{underlined}.}
    \label{tab:map_segmentation}
    \setlength{\tabcolsep}{6pt}
    \renewcommand{\arraystretch}{1.05}
    \small
    \resizebox{0.49\textwidth}{!}{%
    \begin{tabular}{l c c c c c c}
        \toprule
        \multirow{2}{*}{Method} & \multirow{2}{*}{Modality} & \multirow{2}{*}{Pretrain}
        & \multicolumn{4}{c}{IoU $\uparrow$} \\
        \cmidrule(lr){4-7}
        & & & Lane & Drivable & Divider & Crossing \\
        \midrule
        \multirow{2}{*}{UniAD~\cite{hu2023UniAD}}
            & \multirow{2}{*}{C}
            & - & 31.3 & 69.1 & 25.7 & 13.8 \\
            &  & ViDAR~\cite{yang2024vidar} & \underline{35.2} & \textbf{73.9} & \underline{29.1} & \textbf{17.1} \\
        \midrule
        \multirow{2}{*}{UniAD-CRISP}
            & \multirow{2}{*}{C+R}
            & -
            & 32.4 & 69.8 & 26.7 & 15.1 \\
            &
            & \cellcolor{green!5}CRISP
            & \cellcolor{green!5}\textbf{35.4}
            & \cellcolor{green!5}\underline{72.7}
            & \cellcolor{green!5}\textbf{29.3}
            & \cellcolor{green!5}\underline{16.6} \\
        \bottomrule
    \end{tabular}
    }
\end{table}

\begin{table}[t]
    \centering
    \caption{Online tracking results. AMOTA is reported as a percentage, AMOTP in meters, and Recall as a ratio. The best result is shown in \textbf{bold} and the second-best result is \underline{underlined}.}
    \label{tab:tracking}
    \setlength{\tabcolsep}{6pt}
    \renewcommand{\arraystretch}{1.05}
    \small
    \resizebox{0.49\textwidth}{!}{%
    \begin{tabular}{l c c c c c}
        \toprule
        Method & Modality & Pretrain & AMOTA $\uparrow$ & AMOTP $\downarrow$ & Recall $\uparrow$ \\
        \midrule
        \multirow{2}{*}{UniAD~\cite{hu2023UniAD}}
            & \multirow{2}{*}{C}
            & - & 35.9 & 1.32 & 0.48 \\
            &  & ViDAR~\cite{yang2024vidar} & 43.46 & 1.25 & 0.54 \\
        \midrule
        SpaRC-AD~\cite{SpaRC-AD}
            & C+R
            & -
            & \underline{46.9} & \underline{1.13} & 0.55 \\
        \midrule
        \multirow{2}{*}{UniAD-CRISP}
            & \multirow{2}{*}{C+R}
            & -
            & 44.6 & 1.17 & \underline{0.56} \\
            &
            & \cellcolor{green!5}CRISP
            & \cellcolor{green!5}\textbf{48.6}
            & \cellcolor{green!5}\textbf{1.12}
            & \cellcolor{green!5}\textbf{0.57} \\
        \bottomrule
    \end{tabular}
    }
\end{table}

\noindent \textbf{3D Object Detection.}
Table~\ref{tab:main-detection} reports 3D detection results on the nuScenes \emph{val} set. In addition to BEVFormer-based baselines, we include UVTR~\cite{li2022unifying_uvtr}, a unified CO 3D perception model, and its pretrained variants UniPAD~\cite{yang2024unipad} and VisionPAD~\cite{visionpad}. BEVFormer-CRISP achieves 53.2 mAP and 61.3 NDS, outperforming both CO pretraining baselines and the CR SpaRC-AD baseline. Compared with ViDAR-pretrained BEVFormer, the gain suggests that CR forecasting-based pretraining provides a stronger BEV initialization for object-centric 3D perception. The improvement over SpaRC-AD further indicates that the benefit is not simply from adding radar, but from pretraining a reusable CR backbone before detection finetuning.

\noindent \textbf{Online Map Segmentation.}
Table~\ref{tab:map_segmentation} reports online map segmentation results. UniAD-CRISP achieves the best lane and divider IoU, while remaining competitive on drivable area and crossing. The mixed trend is reasonable: map segmentation is dominated by mostly static scene elements, where camera semantics and spatial layout already provide strong cues, and radar contributes less directly than in dynamic-object tasks. Nevertheless, CRISP still improves over CO UniAD across all categories and matches or exceeds ViDAR on the finer lane and divider classes, suggesting that CR pretraining does not harm static BEV structure while primarily benefiting tasks that require motion and geometry.

\noindent \textbf{Online Tracking.}
Table~\ref{tab:tracking} evaluates multi-object tracking. UniAD-CRISP improves AMOTA, AMOTP, and Recall over both CO baselines and SpaRC-AD. This is consistent with the role of radar in the backbone: tracking depends not only on detecting objects in each frame, but also on maintaining stable object states across time. Radar range and motion cues provide complementary evidence for dynamic agents, helping the temporal BEV representation preserve object continuity under motion, occlusion, and appearance ambiguity.

\noindent \textbf{Motion Forecasting.}
Table~\ref{tab:motion_forecasting} reports motion forecasting results. UniAD-CRISP achieves the best minADE, minFDE, and EPA, while remaining second-best on MR. This indicates that CRISP improves the accuracy of the selected trajectory modes and the overall end-to-end prediction score. The result is aligned with the pretraining objective: forecasting future 3D structure encourages the backbone to encode dynamic scene evolution, and radar provides direct motion observations that are useful for downstream agent prediction.

\noindent \textbf{Future Occupancy Prediction.}
Table~\ref{tab:future_occupancy_prediction} evaluates dense future occupancy prediction. UniAD-CRISP achieves the best results across near and far regions for both VPQ and IoU, with especially strong gains in the far field. This is where radar is particularly valuable: CO depth estimates become less reliable at long range, while radar provides metric range measurements and motion-sensitive returns. Since future occupancy prediction requires both spatial geometry and temporal evolution, the combination of LiDAR-supervised forecasting pretraining and radar-aware BEV encoding transfers naturally to this task.

\noindent \textbf{Planning.}
Table~\ref{tab:planning_results} reports planning performance on the nuScenes \emph{val} set. UniAD is explicitly designed as a planning-oriented framework, so improvements here are especially meaningful. UniAD-CRISP reduces both average $\mathcal{L}_2$ error and collision rate, outperforming ViDAR-pretrained UniAD and SpaRC-AD. The lower collision rate suggests that the improved CR BEV representation is not only beneficial for intermediate perception metrics, but also for open-loop safety-related planning metrics. Overall, the downstream results show that CRISP transfers most strongly to tasks where geometry, temporal consistency, and object dynamics are central, while remaining competitive on more static BEV understanding tasks such as map segmentation.

\begin{table}[t]
    \centering
    \caption{Motion forecasting results. minADE and minFDE are reported in meters, while MR and EPA are unitless. The best result is shown in \textbf{bold} and the second-best result is \underline{underlined}.}
    \label{tab:motion_forecasting}
    \setlength{\tabcolsep}{6pt}
    \renewcommand{\arraystretch}{1.05}
    \small
    \resizebox{0.49\textwidth}{!}{%
    \begin{tabular}{l c c c c c c}
        \toprule
        Method & Modality & Pretrain & minADE $\downarrow$ & minFDE $\downarrow$ & MR $\downarrow$ & EPA $\uparrow$ \\
        \midrule
        \multirow{2}{*}{UniAD~\cite{hu2023UniAD}}
            & \multirow{2}{*}{C}
            & - & 0.71 & 1.02 & 0.151 & 0.456 \\
            &  & ViDAR~\cite{yang2024vidar} & 0.65 & 0.97 & 0.147 & 0.513 \\
        \midrule
        SpaRC-AD~\cite{SpaRC-AD}
            & C+R
            & -
            & 0.58 & 0.93 & \textbf{0.121} & 0.530 \\
        \midrule
        \multirow{2}{*}{UniAD-CRISP}
            & \multirow{2}{*}{C+R}
            & -
            & \underline{0.54} & \underline{0.87} & 0.139 & \underline{0.545} \\
            &
            & \cellcolor{green!5}CRISP
            & \cellcolor{green!5}\textbf{0.53}
            & \cellcolor{green!5}\textbf{0.85}
            & \cellcolor{green!5}\underline{0.134}
            & \cellcolor{green!5}\textbf{0.571} \\
        \bottomrule
    \end{tabular}
    }
\end{table}

\begin{table}[t]
    \centering
    \caption{Future occupancy prediction results. All values are percentages. The best result is shown in \textbf{bold} and the second-best result is \underline{underlined}.}
    \label{tab:future_occupancy_prediction}
    \setlength{\tabcolsep}{6pt}
    \renewcommand{\arraystretch}{1.05}
    \small
    \resizebox{0.49\textwidth}{!}{%
    \begin{tabular}{l c c c c c c}
        \toprule
        Method & Modality & Pretrain & VPQ-n $\uparrow$ & VPQ-f $\uparrow$ & IoU-n $\uparrow$ & IoU-f $\uparrow$ \\
        \midrule
        \multirow{2}{*}{UniAD~\cite{hu2023UniAD}}
            & \multirow{2}{*}{C}
            & - & 54.7 & 33.5 & 63.4 & 40.2 \\
            &  & ViDAR~\cite{yang2024vidar} & 58.1 & 37.8 & 65.7 & 42.6 \\
        \midrule
        \multirow{2}{*}{UniAD-CRISP}
            & \multirow{2}{*}{C+R}
            & -
            & \underline{60.8} & \underline{45.5} & \underline{67.7} & \underline{46.6} \\
            &
            & \cellcolor{green!5}CRISP
            & \cellcolor{green!5}\textbf{62.7}
            & \cellcolor{green!5}\textbf{47.5}
            & \cellcolor{green!5}\textbf{69.1}
            & \cellcolor{green!5}\textbf{48.0} \\
        \bottomrule
    \end{tabular}
    }
\end{table}

\begin{table}[t]
    \centering
    \caption{End-to-end planning results. Avg. $\mathcal{L}_2$ is reported in meters, and Avg. Collision is reported as a percentage. The best result is shown in \textbf{bold} and the second-best result is \underline{underlined}.}
    \label{tab:planning_results}
    \setlength{\tabcolsep}{6pt}
    \renewcommand{\arraystretch}{1.05}
    \small
    \resizebox{0.49\textwidth}{!}{%
    \begin{tabular}{l c c c c}
        \toprule
        Method & Modality & Pretrain & Avg. $\mathcal{L}_2$ (m) $\downarrow$ & Avg. Collision (\%) $\downarrow$ \\
        \midrule
        UniAD~\cite{hu2023UniAD}
            & C
            & -
            & 1.03
            & 0.31 \\
        UniAD~\cite{hu2023UniAD}
            & C
            & ViDAR~\cite{yang2024vidar}
            & 0.94
            & 0.20 \\
        \midrule
        SpaRC-AD~\cite{SpaRC-AD}
            & C+R
            & -
            & 0.93
            & 0.23 \\
        \midrule
        \multirow{2}{*}{UniAD-CRISP}
            & \multirow{2}{*}{C+R}
            & -
            & \underline{0.85} & \underline{0.20} \\
            &
            & \cellcolor{green!5}CRISP
            & \cellcolor{green!5}\textbf{0.80}
            & \cellcolor{green!5}\textbf{0.15} \\
        \bottomrule
    \end{tabular}
    }
\end{table}

\subsection{Label-Efficiency of Pretraining}
\label{sec:label_efficiency}

We further evaluate whether the pretrained CRISP backbone improves label efficiency during downstream finetuning. We use the same BEVFormer-CRISP detection architecture and train it for 12 epochs with different fractions of the nuScenes training split: $1/4$, $1/2$, and the full set. We compare two initialization settings: training from scratch and initializing from CRISP pretraining. The scratch baseline still uses the same FCOS3D-pretrained camera encoder as in our main experiments, so the comparison directly highlights the benefit of CRISP backbone pretraining for camera-radar detection.

\begin{figure}[t]
    \centering
\includegraphics[width=1.0\linewidth]{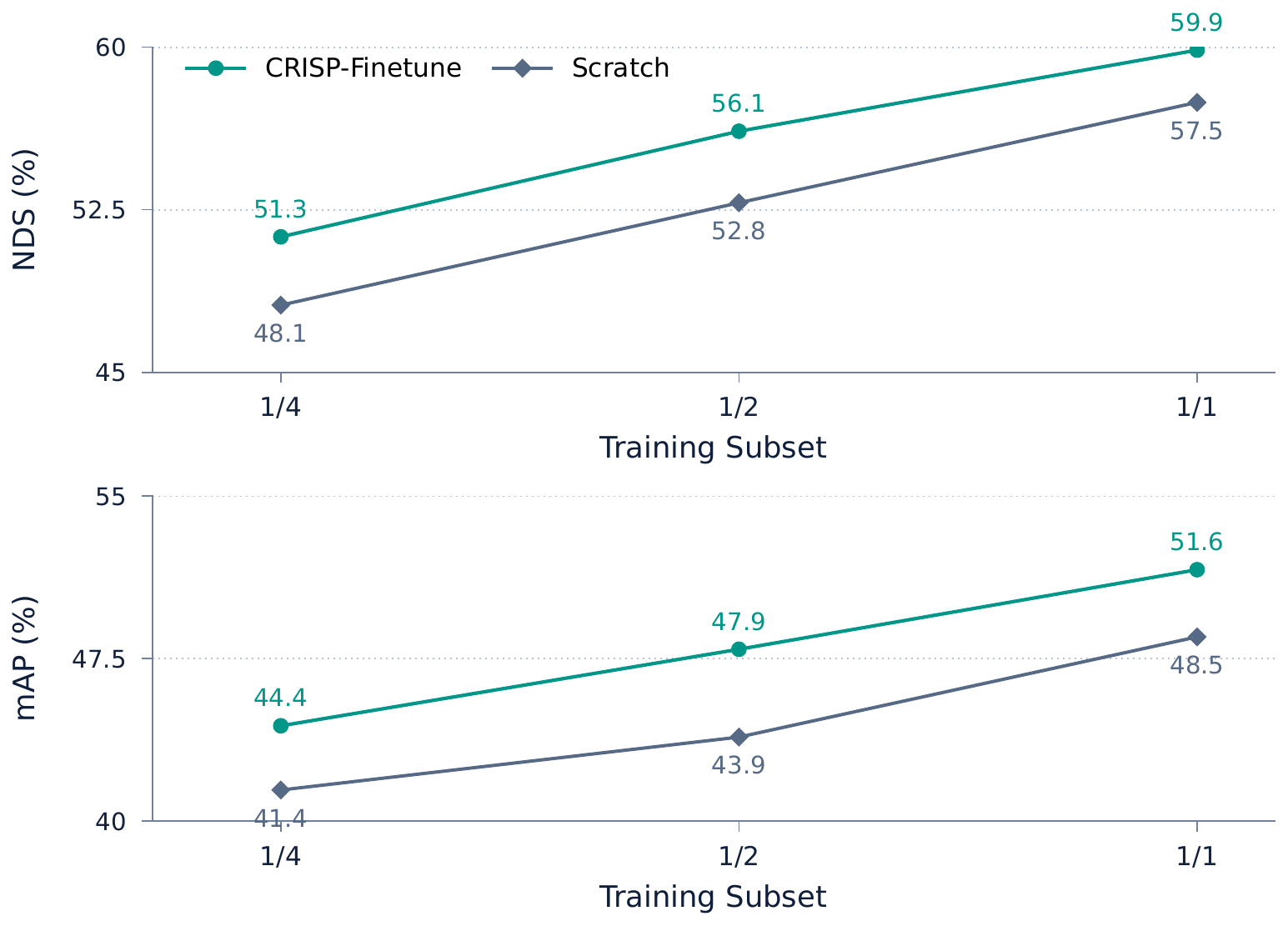}
    \caption{Label-efficiency of CRISP pretraining on nuScenes 3D detection. We train the same BEVFormer-CRISP detection model for 12 epochs using $1/4$, $1/2$, and the full nuScenes training split. The scratch baseline uses the same FCOS3D-pretrained camera encoder but does not use CRISP backbone pretraining. CRISP improves both NDS and mAP across all data scales, with clear gains in the low-label regime.}
    \label{fig:label_efficiency}
    \vspace{-2mm}
\end{figure}

As shown in Fig.~\ref{fig:label_efficiency}, CRISP consistently improves both mAP and NDS across all training-set sizes. The gain is especially clear in the low-label regime: with only $1/4$ or $1/2$ of the training data, CRISP improves NDS and mAP by more than 3 percentage points. The improvement remains stable as more annotations are used, reaching 59.9 NDS and 51.6 mAP with the full training set. These results show that the pretrained CRISP backbone provides a stronger initialization for downstream detection, reducing the dependence on dense 3D box annotations while still improving full-data performance.

\subsection{Ablation Study}

\subsubsection{Implementation Details}
To reduce GPU memory usage, we mainly follow the lightweight BEVFormer-small setting~\cite{li2022bevformer} for ablation experiments. Compared with the full CRISP model, the ablation setting uses a smaller $160\times160$ BEV grid, single-scale image features extracted from images resized to $0.8\times$ of the full input resolution, and a 3-layer BEV encoder, while keeping the training recipe consistent across variants. For efficiency, ablation pretraining uses a fixed $1/8$ subset of the nuScenes \emph{train} split, and detection finetuning uses a fixed $1/4$ subset. All ablation models are trained with the same loss configuration as the full model, and still evaluated on the full nuScenes validation split. We train all ablation models using 4 GPUs.

The ablations are designed to assess both the contribution of each CRISP component and the effectiveness of our specific design choices in the radar encoder, radar-enhanced temporal self-attention, and multimodal feature rendering. Unless otherwise noted, we evaluate forecasting with Chamfer Distance (CD) and Average Euclidean Error (AEE), and evaluate downstream detection with mean Average Precision (mAP), nuScenes Detection Score (NDS), and mean Average Velocity Error (mAVE). CD measures set-level similarity between predicted and ground-truth point clouds using squared nearest-neighbor distances, while AEE measures metric prediction accuracy in meters after associating predictions with corresponding ground-truth viewing directions. For detection, mAP and NDS capture overall object detection quality, while mAVE measures velocity estimation error in meters per second, which is closely related to dynamic scene understanding.

\subsubsection{Component Ablation}

Table~\ref{tab:ablation_main} studies the cumulative contribution of the main CRISP components. Adding radar-enhanced temporal self-attention gives a large forecasting gain over ViDAR, confirming that radar motion cues are useful when injected before temporal aggregation. However, downstream transfer remains incomplete: mAVE is still high, suggesting that radar-aware temporal propagation alone does not fully provide motion-consistent BEV features for detection. Replacing ViDAR's camera-centric latent rendering with multimodal feature rendering addresses this issue. As discussed in Section~\ref{sec:method-multimodal-rendering}, latent rendering is suitable when geometry must be inferred from images, but in CRISP radar already provides calibrated metric BEV features with range and motion cues; therefore, camera and radar evidence should be composed directly inside the BEV backbone. This change produces the largest downstream improvement, especially in NDS and mAVE. The enhanced radar encoder then gives a smaller but consistent final gain, validating the full design of ego-aware radar encoding, radar-enhanced temporal propagation, and iterative multimodal evidence rendering.

\subsubsection{Radar Encoder Design}

Table~\ref{tab:ablation_radar_encoder} ablates the enhanced radar encoder. The PointPillars baseline already provides a reasonable radar BEV feature, but it does not explicitly account for the ego-centric nature of radar motion measurements. Adding ego-motion awareness improves AEE and slightly improves NDS, suggesting that ego-conditioned radar features help interpret Doppler cues more consistently; however, mAP slightly drops and mAVE worsens, so the effect is not uniformly positive for detection. Residual spatial calibration then improves CD and gives the best or tied-best detection metrics, indicating that aligning radar BEV features with the camera-side BEV representation is useful before radar participates in temporal attention and multimodal rendering.

\begin{table}[t]
    \centering
    \caption{Main ablation study of the proposed CRISP modules. Starting from the ViDAR baseline, each subsequent row cumulatively adds one additional component on top of the previous row: radar-enhanced TSA, multimodal feature rendering, and the enhanced radar encoder. mAP and NDS are reported in percentage points (\%), and mAVE is reported in meters per second (m/s). The best result for each metric is \textbf{bolded}.}
    \label{tab:ablation_main}
    \setlength{\tabcolsep}{5pt}
    \renewcommand{\arraystretch}{1.15}
    \resizebox{0.5\textwidth}{!}{%
    \begin{tabular}{l ccc ccc ccc}
        \toprule
        \multirow{2}{*}{Variant}
        & \multicolumn{3}{c}{CD ($\mathrm{m}^2$) $\downarrow$}
        & \multicolumn{3}{c}{AEE ($\mathrm{m}$) $\downarrow$}
        & \multicolumn{3}{c}{Detection Finetuning} \\
        \cmidrule(lr){2-4} \cmidrule(lr){5-7} \cmidrule(lr){8-10}
        & 1s & 2s & 3s & 1s & 2s & 3s & mAP $\uparrow$ & NDS $\uparrow$ & mAVE $\downarrow$ \\
        \midrule
        ViDAR~\cite{yang2024vidar} & 1.42 & 1.65 & 1.95 & 2.67 & 2.81 & 2.98 & 31.3 & 34.9 & 1.19 \\
        + radar-enhanced TSA & 0.97 & 1.22 & 1.54 & 1.78 & 1.93 & 2.14 & 37.8 & 39.8 & 1.23 \\
        + multimodal feature rendering & 0.90 & 1.12 & 1.43 & 1.68 & 1.83 & 2.01 & 41.1 & 48.8 & \textbf{0.49} \\
        + enhanced radar encoder & \textbf{0.88} & \textbf{1.10} & \textbf{1.41} & \textbf{1.66} & \textbf{1.78} & \textbf{1.93} & \textbf{41.8} & \textbf{49.5} & \textbf{0.49} \\
        \bottomrule
    \end{tabular}%
    }
\end{table}

\begin{table}[t]
    \centering
    \caption{Ablation study of the radar encoder design. Starting from the PointPillars radar encoder, each subsequent row cumulatively adds one additional component on top of the previous row: ego-motion awareness and residual spatial calibration. All other CRISP modules are kept the same for fair comparison. mAP and NDS are reported in percentage points (\%), and mAVE is reported in meters per second (m/s). The best result for each metric is \textbf{bolded}.}
    \label{tab:ablation_radar_encoder}
    \setlength{\tabcolsep}{5pt}
    \renewcommand{\arraystretch}{1.15}
    \resizebox{0.5\textwidth}{!}{%
    \begin{tabular}{l ccc ccc ccc}
        \toprule
        \multirow{2}{*}{Variant}
        & \multicolumn{3}{c}{CD ($\mathrm{m}^2$) $\downarrow$}
        & \multicolumn{3}{c}{AEE ($\mathrm{m}$) $\downarrow$}
        & \multicolumn{3}{c}{Detection Finetuning} \\
        \cmidrule(lr){2-4} \cmidrule(lr){5-7} \cmidrule(lr){8-10}
        & 1s & 2s & 3s & 1s & 2s & 3s & mAP $\uparrow$ & NDS $\uparrow$ & mAVE $\downarrow$ \\
        \midrule
        PointPillars~\cite{lang2019pointpillars}
        & 0.90 & 1.12 & 1.43
        & 1.68 & 1.82 & 2.01
        & 41.1 & 48.8 & \textbf{0.49} \\
        + Ego-motion awareness
        & 0.90 & 1.12 & 1.43
        & \textbf{1.66} & \textbf{1.78} & \textbf{1.92}
        & 41.0 & 49.0 & 0.51 \\
        + Residual spatial calibration
        & \textbf{0.88} & \textbf{1.10} & \textbf{1.41}
        & \textbf{1.66} & \textbf{1.78} & 1.93
        & \textbf{41.8} & \textbf{49.5} & \textbf{0.49} \\
        \bottomrule
    \end{tabular}%
    }
\end{table}

\subsubsection{Radar-Enhanced Temporal Self-Attention}

Table~\ref{tab:ablation_tsa} evaluates the two design choices in radar-enhanced temporal self-attention. Gated radar conditioning mostly improves over standard TSA, especially on AEE and detection metrics, although CD at 3s slightly worsens. This shows that radar-primed BEV queries make temporal propagation more effective for many, but not all, metrics. Adding temporal gating gives the strongest aggregate result, especially on AEE and downstream velocity estimation. This supports the view that temporal aggregation should not uniformly average current and historical branches; instead, each BEV token should decide how much current evidence and propagated memory to use.

\subsubsection{Multimodal Feature Rendering}

Table~\ref{tab:ablation_rendering} compares different ways to compose camera and radar evidence after temporal aggregation. Adaptive scalar weighting is competitive for forecasting, including the best 3s CD, but weaker for downstream transfer, while full gating degrades detection performance, suggesting that gating complete modality features can disturb the existing BEV state. Innovation gating achieves the best downstream transfer while maintaining competitive forecasting because it gates only the modality-specific residual updates proposed by camera and radar cross-attention. This preserves the current BEV representation while allowing each layer to selectively admit useful sensor evidence, which is the intended behavior of the MIG.

Overall, the ablations show that CRISP's improvements come from introducing radar at multiple stages of the BEV state update process. Radar-enhanced TSA improves temporal propagation, multimodal feature rendering provides the main transfer gain by replacing camera-centric latent rendering, and the enhanced radar encoder improves the quality of the radar evidence used throughout the backbone.

\begin{table}[t]
    \centering
    \caption{Ablation study of design choices in radar-enhanced temporal self-attention. Starting from standard TSA, each subsequent row cumulatively adds gated radar conditioning and temporal gating. All other CRISP modules are kept the same for fair comparison. mAP and NDS are reported in percentage points (\%), and mAVE is reported in meters per second (m/s). The best result for each metric is \textbf{bolded}.}
    \label{tab:ablation_tsa}
    \setlength{\tabcolsep}{5pt}
    \renewcommand{\arraystretch}{1.15}
    \resizebox{0.5\textwidth}{!}{%
    \begin{tabular}{l ccc ccc ccc}
        \toprule
        \multirow{2}{*}{Variant}
        & \multicolumn{3}{c}{CD ($\mathrm{m}^2$) $\downarrow$}
        & \multicolumn{3}{c}{AEE ($\mathrm{m}$) $\downarrow$}
        & \multicolumn{3}{c}{Detection Finetuning} \\
        \cmidrule(lr){2-4} \cmidrule(lr){5-7} \cmidrule(lr){8-10}
        & 1s & 2s & 3s & 1s & 2s & 3s & mAP $\uparrow$ & NDS $\uparrow$ & mAVE $\downarrow$ \\
        \midrule
        Standard TSA
        & 0.91 & 1.12 & \textbf{1.41}
        & 1.69 & 1.82 & 1.99
        & 40.8 & 48.6 & 0.52 \\
        + Gated radar conditioning
        & 0.89 & 1.12 & 1.43
        & 1.67 & 1.80 & 1.95
        & 41.7 & 49.4 & 0.52 \\
        + Temporal Gating
        & \textbf{0.88} & \textbf{1.10} & \textbf{1.41}
        & \textbf{1.66} & \textbf{1.78} & \textbf{1.93}
        & \textbf{41.8} & \textbf{49.5} & \textbf{0.49} \\
        \bottomrule
    \end{tabular}%
    }
\end{table}

\begin{table}[t]
    \centering
    \caption{Ablation study of modality fusion design in multimodal feature rendering. All other CRISP modules are kept the same for fair comparison. mAP and NDS are reported in percentage points (\%), and mAVE is reported in meters per second (m/s). The best result for each metric is \textbf{bolded}.}
    \label{tab:ablation_rendering}
    \setlength{\tabcolsep}{5pt}
    \renewcommand{\arraystretch}{1.15}
    \resizebox{0.5\textwidth}{!}{%
    \begin{tabular}{l ccc ccc ccc}
        \toprule
        \multirow{2}{*}{Variant}
        & \multicolumn{3}{c}{CD ($\mathrm{m}^2$) $\downarrow$}
        & \multicolumn{3}{c}{AEE ($\mathrm{m}$) $\downarrow$}
        & \multicolumn{3}{c}{Detection Finetuning} \\
        \cmidrule(lr){2-4} \cmidrule(lr){5-7} \cmidrule(lr){8-10}
        & 1s & 2s & 3s & 1s & 2s & 3s & mAP $\uparrow$ & NDS $\uparrow$ & mAVE $\downarrow$ \\
        \midrule
        Adaptive scalar weight~\cite{li2024bevformer_tpami}
        & 0.90 & 1.12 & \textbf{1.40}
        & 1.67 & 1.79 & 1.95
        & 41.2 & 48.9 & 0.53 \\
        Full Gating
        & 0.90 & 1.12 & 1.46
        & 1.67 & 1.79 & 1.95
        & 40.5 & 47.7 & 0.57 \\
        Innovation Gating
        & \textbf{0.88} & \textbf{1.10} & 1.41
        & \textbf{1.66} & \textbf{1.78} & \textbf{1.93}
        & \textbf{41.8} & \textbf{49.5} & \textbf{0.49} \\
        \bottomrule
    \end{tabular}%
    }
\end{table}

\begin{figure*}[h]
    \centering
    \includegraphics[width=1.0\linewidth]{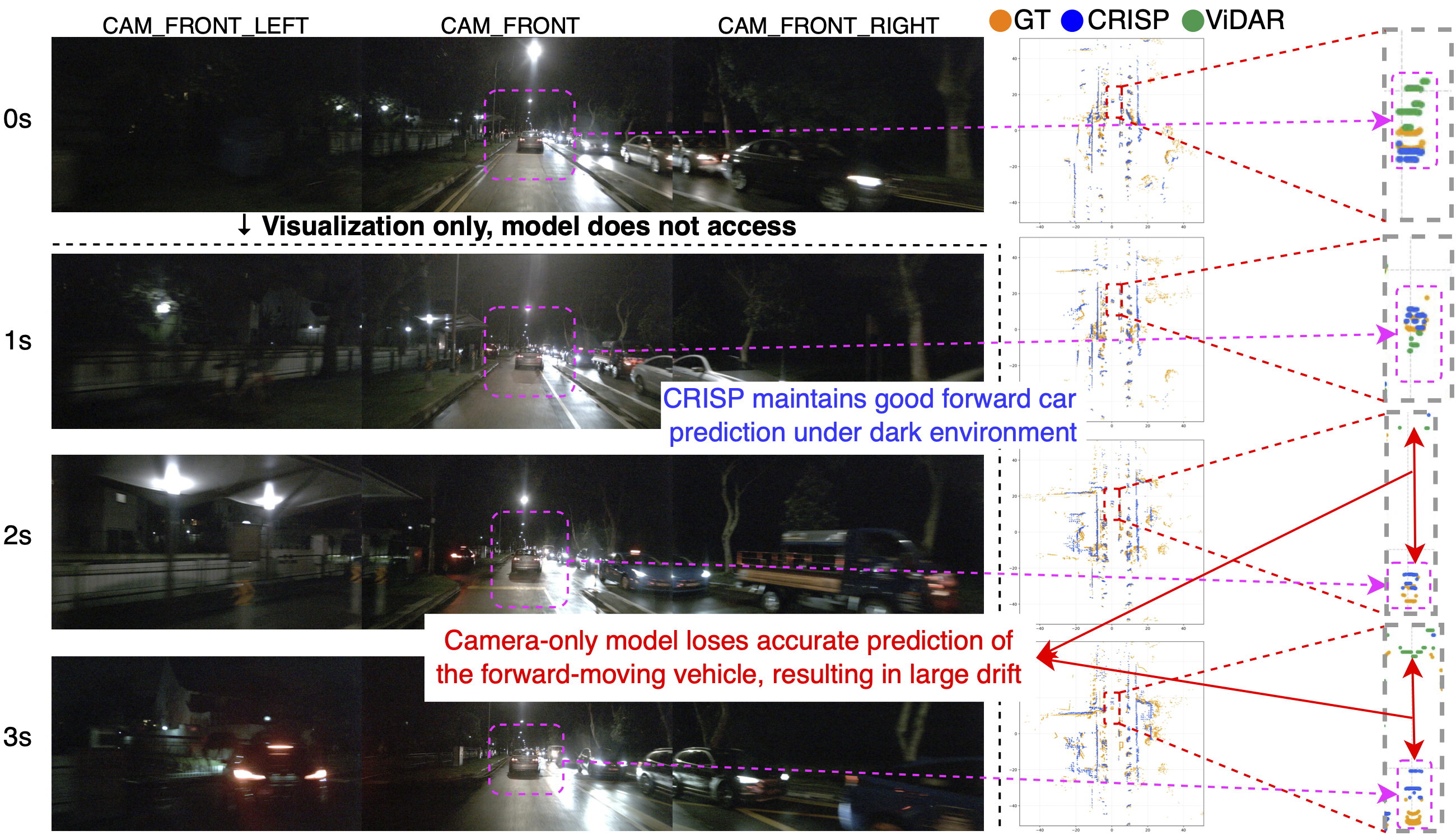}
   \caption{Qualitative comparison between CRISP and ViDAR for future world-geometry prediction in a low-light scene. We show the front camera views at the current frame (0s) and future timestamps for visualization only; future images are not provided to the model. On the right, predicted point clouds are compared with ground truth. In this example, the CO ViDAR baseline loses the forward vehicle over time and exhibits large spatial drift, while CRISP maintains a more accurate prediction of the vehicle trajectory. This illustrates the benefit of radar motion cues for forecasting dynamic objects under challenging illumination, which is important for downstream trajectory evaluation and safe planning around leading vehicles.}
    \label{fig:qualitative_good_2}
    \vspace{-2mm}
\end{figure*}

\begin{figure*}[h]
    \centering
    \includegraphics[width=1.0\linewidth]{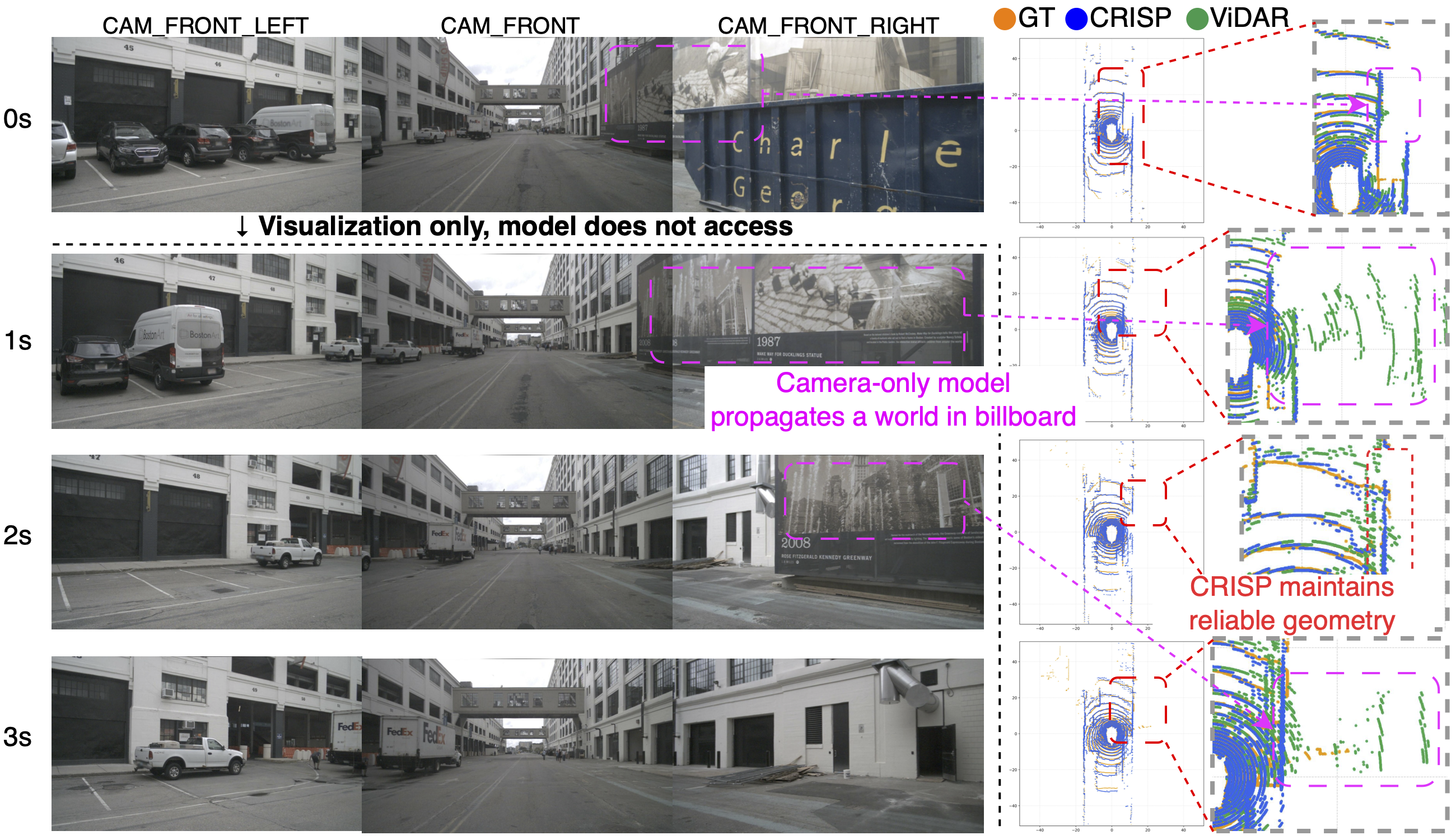}
   \caption{Qualitative comparison between CRISP and ViDAR for future world-geometry prediction. We show the front camera views at the current frame (0s) and future timestamps for visualization only; future images are not provided to the model. On the right, predicted point clouds are compared with ground truth, where CRISP better preserves the scene geometry over time. In this example, the CO ViDAR baseline propagates an erroneous structure from a billboard region, while CRISP maintains more reliable geometry.}
    \label{fig:qualitative_good_1}
    \vspace{-2mm}
\end{figure*}

\begin{figure*}[t]
    \centering
    \includegraphics[width=1.0\linewidth]{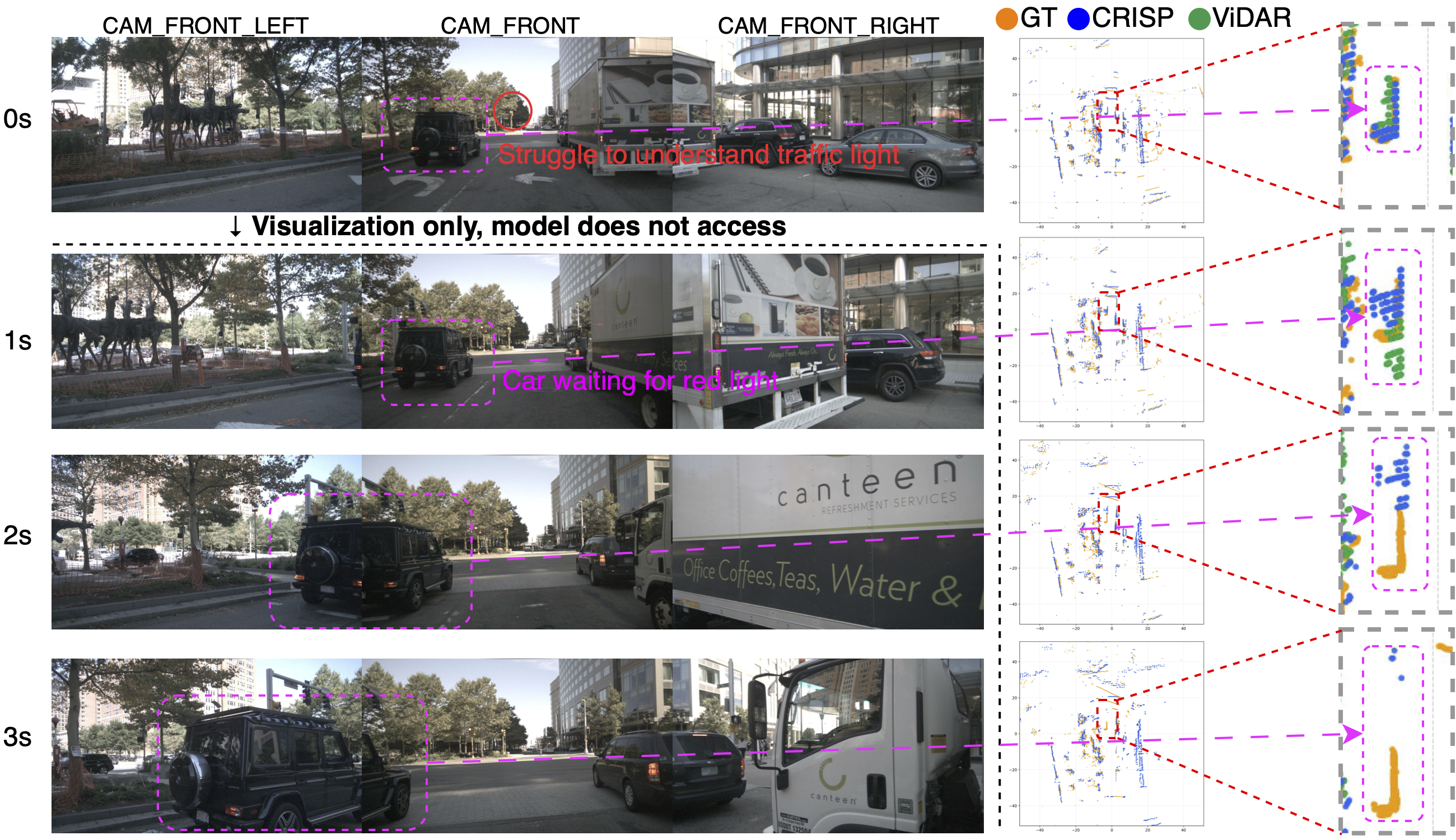}
   \caption{Failure case analysis for future world-geometry prediction. We show the front camera views at the current frame (0s) and future timestamps for visualization only; future images are not provided to the model. On the right, predicted point clouds are compared with ground truth. This scene requires reasoning about a traffic light that is small and difficult to observe from the input images: the highlighted vehicle remains stopped for the red light, while the road ahead is visually open. CRISP predicts the vehicle geometry more accurately than the CO ViDAR baseline at early horizons, but its prediction begins to drift after 2s. This suggests that, when high-level traffic-signal intent is ambiguous, the forecasting model may rely more on learned motion priors from pretraining than on the underlying traffic-control state.}
    \label{fig:qualitative_failure}
    \vspace{-2mm}
\end{figure*}

\subsection{Qualitative Analysis}
\label{sec:qualitative_analysis}

We visualize future point cloud prediction results to better understand what CRISP learns from CR forecasting pretraining. In all examples, the model only observes historical CR inputs; future camera frames are shown only for visualization. Predicted point clouds are compared with ground truth over future horizons, with CRISP and ViDAR shown side by side.

Figures~\ref{fig:qualitative_good_2} and~\ref{fig:qualitative_good_1} illustrate two representative strengths of CRISP. In the low-light scene of Fig.~\ref{fig:qualitative_good_2}, the CO ViDAR baseline loses the forward vehicle over time and exhibits large spatial drift, while CRISP maintains a more stable prediction of the vehicle trajectory. This case highlights the value of radar motion cues under challenging illumination, where visual appearance alone is unreliable. In Fig.~\ref{fig:qualitative_good_1}, ViDAR propagates an erroneous structure from a billboard-like image region into the predicted geometry. CRISP better preserves the physical scene layout, suggesting that radar-enhanced BEV features help anchor the future prediction to metric scene structure rather than purely visual texture.

Figure~\ref{fig:qualitative_failure} shows a remaining failure mode. The highlighted vehicle is stopped at a red light, but the traffic signal is small and difficult to infer from the input views. Although CRISP predicts the vehicle geometry more accurately than ViDAR at early horizons, its prediction begins to drift after 2s. This suggests that when future behavior depends on high-level traffic semantics or agent intent, geometry and motion priors alone may be insufficient. These examples indicate that CRISP improves future world-geometry prediction in visually ambiguous and motion-heavy scenes, while also motivating future work on incorporating stronger traffic-rule, signal-state, or intent-level reasoning.

\section{Discussion}

CRISP frames CR fusion as predictive representation learning rather than task-specific supervised fusion. By forecasting future LiDAR point clouds from historical CR observations, the model learns a reusable spatiotemporal BEV representation that encodes geometry, motion, and cross-modal interactions, while still requiring only camera and radar at inference time. This connects two previously separate directions: forecasting-based world-model pretraining for driving and practical CR fusion for deployable autonomy. The downstream results further suggest that the learned representation is broadly useful beyond the pretext task, with the strongest gains appearing in tasks that depend on geometry, temporal consistency, and agent motion, such as detection, tracking, occupancy prediction, motion forecasting, and planning. The more modest gains on mostly static map elements also suggest that radar contributes most when dynamic scene evolution is important.

Despite these benefits, CRISP still has important failure modes. First, the forecasting objective is data-driven and does not explicitly model causality, traffic rules, or agent intent (Fig.~\ref{fig:qualitative_failure}). In common situations, predicting likely future scene structure is useful; however, in rare safety-critical cases, a model may favor statistically common continuations from the training data even when they are inappropriate for the current scene. For example, near an occluded intersection with a red traffic light, the model may still predict forward traffic flow if similar visual-geometric observations in the training set usually contain moving vehicles. Robust prediction therefore requires higher-level understanding of signals, right-of-way, occlusion, map context, and interactive intent, which are not directly captured by the current objective.

Second, CRISP inherits the limitations of radar sensing. Radar provides valuable range and Doppler cues, but it is sparse, noisy, semantically weak, and vulnerable to clutter, multipath, small objects, static objects, calibration errors, and sensor degradation. The proposed gates can reduce unhelpful modality updates, but they do not guarantee calibrated uncertainty or reliable behavior when radar evidence is missing or contradictory. Future CR backbones should therefore reason not only about how to use radar, but also about when radar should be trusted.

Third, the current study is limited by dataset scale and diversity. CRISP is evaluated on nuScenes, one of the few public datasets with synchronized camera, radar, and LiDAR, but it remains far smaller and less diverse than the scale needed for driving foundation models. Cross-dataset pretraining is a natural next step. Large planning datasets such as nuPlan~\cite{caesar2021nuplan} illustrate the value of scale, but the community still lacks comparably large public datasets with synchronized radar streams. We hope this motivates future releases that include radar together with camera, LiDAR, ego-motion, maps, and planning annotations, enabling CR pretraining at foundation-model scale.

Several research directions follow from these limitations. One is to transfer CRISP-style pretraining to broader architectures, including other BEV encoders, occupancy networks, end-to-end planning systems, and diffusion-based world models. This would help distinguish architecture-specific gains from the general value of CR predictive pretraining. Another direction is to move from deterministic geometric forecasting toward uncertainty-aware and semantics-aware world modeling. Future objectives could combine future geometry with map context, traffic-light states, agent intent, rule annotations, language, and counterfactual ego trajectories, allowing the model to represent multiple plausible futures instead of a single likely one. Such uncertainty is especially important in occluded intersections, unprotected turns, pedestrian interactions, and other safety-critical scenarios.

More broadly, CRISP suggests a useful pattern for robotics: learn deployment-friendly sensor representations from richer training-only supervision. CR driving is one instance, where LiDAR provides pretraining supervision but is removed at inference time. Similar ideas could benefit mobile robots, drones, field robots, and embodied agents that must operate with practical sensor suites under partial observability, distribution shift, sensor degradation, and long-horizon decision making.

\section{Conclusion}

We presented CRISP, a forecasting-based pretraining framework for transferable camera-radar representation learning in autonomous driving. CRISP uses historical multi-view images and radar sweeps to predict future LiDAR point clouds, using LiDAR only as privileged supervision during pretraining. After pretraining, the forecasting decoder is removed, leaving a practical LiDAR-free CR backbone for downstream inference. By moving CR interaction inside the spatiotemporal BEV encoder, CRISP allows radar cues to guide temporal propagation and multimodal BEV feature formation rather than serving only as a late-fusion signal. CRISP combines an enhanced radar encoder, radar-enhanced temporal self-attention, and multimodal feature rendering with modality innovation gating to jointly encode visual semantics, radar motion cues, BEV geometry, and temporal dynamics. Experiments on nuScenes show that CRISP improves future point cloud forecasting and transfers broadly to 3D detection, tracking, online mapping, motion forecasting, future occupancy prediction, and planning, with especially strong gains on tasks requiring geometry and dynamic scene reasoning. Overall, CRISP shows that CR fusion can be studied as a scalable predictive pretraining problem, and we hope it encourages larger-scale CR pretraining, broader backbone transfer, and semantics- and uncertainty-aware driving world models.

\bibliographystyle{IEEEtran}
\bibliography{references}

\end{document}